\documentclass[letterpaper, 10 pt, conference]{ieeeconf}  % Comment this line out if you need a4paper

\IEEEoverridecommandlockouts                              % This command is only needed if 
\usepackage{cite}
\usepackage{amsmath,amssymb,amsfonts}
\usepackage[belowskip=-10pt]{caption}
\usepackage{booktabs}
\usepackage{algorithmic}
\usepackage{graphicx}
\usepackage{subcaption}
\usepackage{hyperref}
\usepackage{textcomp}
\usepackage{xcolor}
\usepackage{float}
\usepackage{makecell}
\title{\LARGE \bf
EpiTransfer: Sparse, Training-Free Long-Range Depth Estimation from Temporal Monocular Aerial Frames
}

\author{Diksha Aggarwal$^{1,\dag}$, Rutvik Dagadkhair$^{1}$, Sanjana Srivastava$^{2}$,\\ Bradley Denby$^{2}$ and Kevin Kochersberger$^{1}$% <-this % stops a space
\thanks{*This work was not supported by any organization}% <-this % stops a space
\thanks{$^{1}$Department of Mechanical Engineering, Virginia Tech, Blacksburg, Virginia, USA.}
\thanks{$^{2}$Department of Aerospace and Ocean Engineering, Virginia Tech, Blacksburg, Virginia.}
\thanks{$^{\dag}${Corresponding Author: \tt\small diksha@vt.edu}}
}

\begin{document}

\maketitle
\thispagestyle{empty}
\pagestyle{empty}

%%%%%%%%%%%%%%%%%%%%%%%%%%%%%%%%%%%%%%%%%%%%%%%%%%%%%%%%%%%%%%%%%%%%%%%%%%%%%%%%
\begin{abstract}
Reliable 3D spatial understanding is essential for autonomous
navigation, obstacle avoidance, and scene reconstruction. While
state-of-the-art learned depth estimation techniques achieve high
accuracy in-distribution, they often generalize poorly to novel
viewpoints and altitudes. This paper presents a geometrically
derived, training-free depth estimation method using epipolar
transfer with only two monocular images and camera pose estimates.
By leveraging camera motion to synthesize a virtual stereo pair with
a freely chosen baseline, our approach transforms temporal
correspondence into a stereo triangulation task while mitigating
geometric degeneracies inherent to direct two-view triangulation.
Validated across outdoor drone flights (to a maximum range of
approximately 90\,m) and indoor OptiTrack environments against
LiDAR ground truth, the method achieves an indoor AbsRel of 0.092
and $\delta < 1.25$ of 0.940, comparable to direct triangulation
(AbsRel 0.073) while retaining valid depth over a larger fraction
of challenging scenes, and substantially outperforms off-the-shelf learning-based
baselines such as ZoeDepth (AbsRel 0.225) and Depth Anything V2
(AbsRel 0.570), which are not trained or fine-tuned for this domain,
with no training data required.
\end{abstract}

%%%%%%%%%%%%%%%%%%%%%%%%%%%%%%%%%%%%%%%%%%%%%%%%%%%%%%%%%%%%%%%%%%%%%%%%%%%%%%%%
\section{INTRODUCTION}
\label{sec:intro}
Spatial perception provides essential geometric context for mobile robots, supporting fundamental tasks such as mapping, local navigation, and collision avoidance for better robot control. For platforms such as aerial vehicles, where size, weight, power, and cost constraints prevent the use of heavy active sensors like LiDAR, radar, and multi-camera rigs, passive stereo vision provides an attractive alternative by estimating depth through triangulation of corresponding features in synchronized, calibrated camera views \cite{hartley2003multiple}.  However, the fixed physical baseline imposes a fundamental trade-off between compact sensor packaging and long-range depth accuracy. Short-baseline systems produce small disparities and increasingly uncertain depth estimates at long range, while wider baselines increase platform footprint, calibration demands, and occlusion-related correspondence failures \cite{pinggera2014know}.

Monocular cameras are advantageous for such platforms because they are lightweight, low-power, inexpensive, and information-rich. However, their use for geometric perception remains challenging: a single image does not, by itself, determine metric range due to the inherent scale ambiguity of monocular projection \cite{hartley2003multiple,saxena2009make3d}. Moreover, reliable depth recovery in robotic environments must account for image noise, imperfect camera-pose estimates, limited computational resources, and changes in scene or imaging conditions. A practical depth-sensing approach should therefore provide metric estimates efficiently while also quantifying the reliability of individual measurements \cite{poggi2020uncertainty,kendall2017uncertainties}.

For aerial approach and navigation, camera orientation influences the scene geometry available for perception. Nadir-looking cameras are effective for inspecting horizontal ground or rooftop surfaces near a landing location, but provide limited visibility of the forward approach corridor and vertical structures. In contrast, oblique-looking cameras instead observe both ground context and elevated scene elements, including façades, roof edges, walls, poles, towers, and cranes \cite{vacca2017nadir,rossi2017nadir}. Such views are particularly informative in urban environments, where vertical and inclined surfaces not adequately represented in nadir imagery alone. However, the large depth variation, viewpoint-dependent occlusion, and changing scale in oblique imagery make reliable metric depth estimation challenging \cite{qin2015facade,zhang2018integration}.
\begin{figure}
    \centering
    \includegraphics[width=\linewidth]{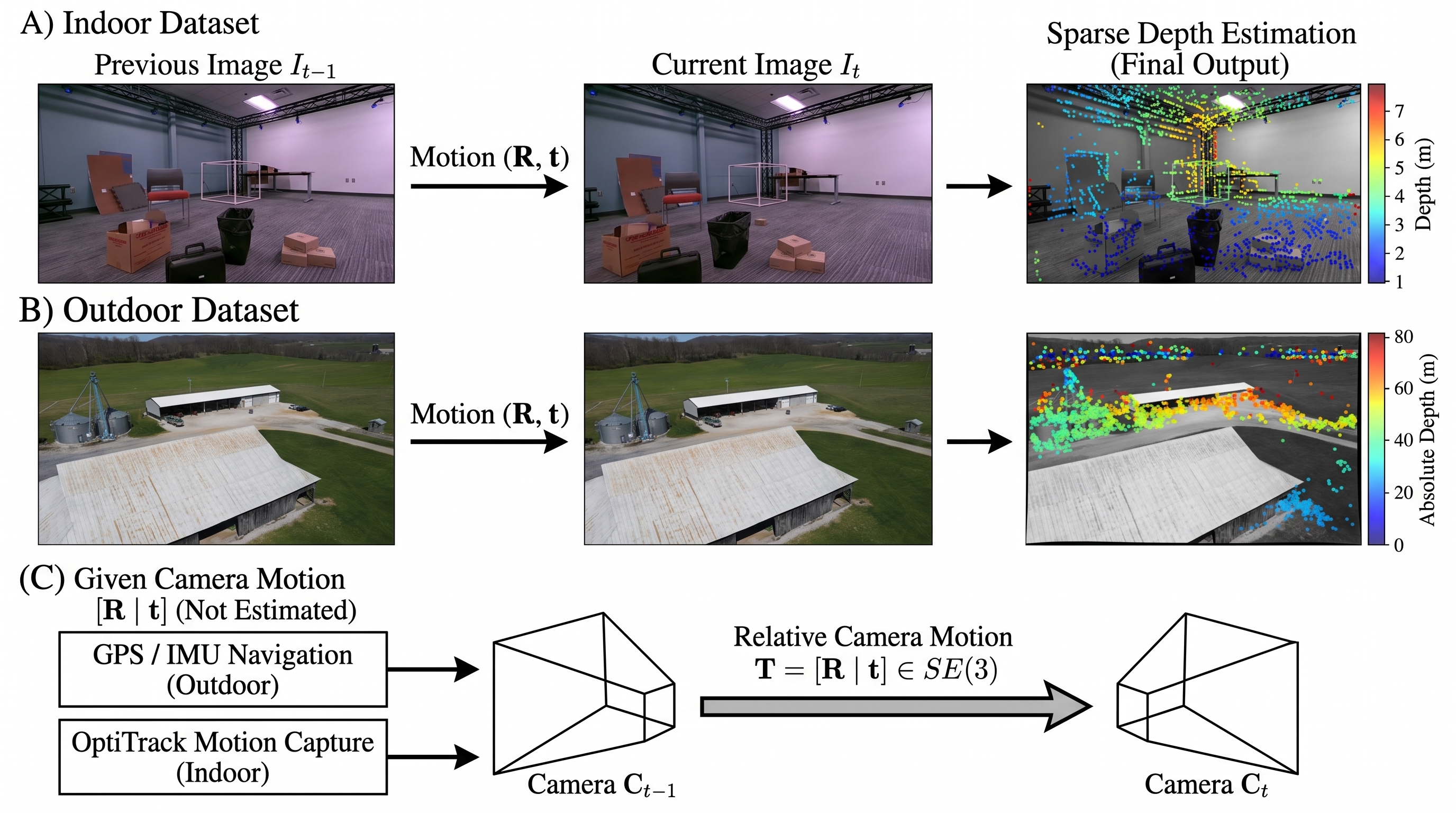}
    \caption{We use temporal monocular images and relative pose of cameras, perform sparse feature matching using XFeat, synthesize a virtual stereo pair via epipolar transfer, and triangulate to obtain sparse depth.}
    \label{fig:overall_method}
\end{figure}

Monocular depth estimation is commonly addressed through
multi-view geometric reconstruction, model-based state
estimation and sensor fusion, or learning-based depth
prediction. Multi-view methods exploit observations from
distinct viewpoints and calibrated camera geometry to recover
scene structure through feature-based, direct, or hybrid
image alignment~\cite{hartley2003multiple,mur2015orb,engel2014lsd,forster2014svo}.
Model-based estimation frameworks combine visual observations
with motion, inertial, or other sensor measurements to
estimate states and maintain geometric representations under
uncertainty~\cite{bloesch2015robocentric,leutenegger2015okvis,campos2021orbslam3}. Learning-based approaches instead train neural networks, using
supervised, self-supervised, or a combination of objectives, to
predict dense depth directly from a monocular image, video
sequence, or sparse-depth input ~\cite{eigen2014depth,godard2019digging,ranftl2021vision,ma2019selfsupervised}.
More recently, transformer- and foundation-model-based methods
leverage large-scale pretrained visual priors to achieve strong
zero-shot, cross-scene generalization, optionally exploiting
sparse depth measurements for dense
completion~\cite{chen2025propagating,cheng2018depth,yan2022rignet}.
These methods provide complementary capabilities, but their
reliability depends on how well the deployment environment
matches the visual, geometric, and sensor characteristics
represented during training, and, being limited by their
training data, they may produce unbounded results under
distribution shift~\cite{guizilini20203dpackingselfsupervisedmonocular,yu2024dme}
%~\cite{poggi2020uncertainty,guizilini20203dpackingselfsupervisedmonocular,yu2024dme}.
A sparse geometric measurement source with
physically-interpretable error characteristics can therefore
complement learned depth, either as input to depth
completion~\cite{ma2019selfsupervised}, a fusion cue, a
validation signal, or a source of uncertainty-aware
constraints for downstream planning.

In this paper, we propose a geometrically derived sparse
depth estimation method --- obtained via epipolar transfer
using only two images and available pose estimates from
sensors like GPS, IMU, or wheel odometry as shown in Fig. \ref{fig:overall_method}. Our key insight
is that camera motion can be leveraged to synthesize a
virtual stereo pair through epipolar transfer, enabling depth
triangulation without a physical stereo rig or multi-frame
optimization. Unlike direct per-pixel epipolar search on the
real second view~\cite{engel2014lsd} or dense learned
rectification into a synthetic view~\cite{brousseau2025spherical},
our approach transfers sparse feature correspondences into a
synthetic third view via closed-form trifocal intersection,
with a baseline decoupled from the real inter-frame motion.
This method can serve as a viable and lightweight alternative
source of sparse depth for such pipelines.

The proposed method, EpiTransfer, is implemented on the images collected in the outdoor and indoor environments. Validation is performed on the indoor images using a LiDAR. Experimental results demonstrate competitive accuracy with significantly lower errors in comparison to the state-of-the-art depth estimation methods. 

The main contributions of this paper are:
\begin{itemize}
    \item A novel two-frame depth estimation formulation that
    synthesizes a virtual stereo pair via epipolar transfer
    using only camera motion, validated to a maximum range of
    approximately 90\,m outdoors --- roughly $3\times$ the range
    reported by prior geometric approaches to aerial approach
    and landing.
    \item A hybrid pipeline combining deep learning-based
    feature correspondences with training-free geometric
    triangulation for robust depth estimation.
    \item An explicit, geometrically derived characterization of the method's failure mode near the trifocal plane, enabling principled masking of ill-conditioned depth estimates without empirical thresholding.
\end{itemize}

The remainder of this paper is organized as follows: Section~II
reviews related work in geometric, learning-based, and
visual-inertial depth estimation. Section~III details the
proposed depth estimation methodology. Section~IV describes
the experimental setup, including the hardware platform,
datasets, and evaluation protocol. Section~V presents the
experimental results and comparative analysis, and Section~VI
concludes the paper.

\section{Related Work}

Depth estimation from RGB cameras has been studied along stereo, multi-view geometric, and learning-based monocular lines. This section situates EpiTransfer among these, and in particular among prior methods that turn two frames of a single moving camera into a stereo-like pair.

\subsection{Geometric Two- and Multi-View Depth from Monocular Motion}
Classical stereo recovers depth from disparities between calibrated,
fixed-baseline views using block matching or Semi-Global Matching,
and recent deep stereo networks such as PSMNet \cite{PSM} and RAFT-Stereo \cite{RAFT-Stereo}
improve matching robustness in low-texture regions; however, all of
these methods require a physical rig, and their achievable depth
range and precision are fundamentally limited by the fixed physical
baseline~\cite{pinggera2014know}. Multi-view SfM pipelines such as
COLMAP~\cite{schoenberger2016sfm} jointly recover poses and structure
through global bundle adjustment, yielding geometrically consistent
but computationally expensive, typically offline reconstructions from
many views. Visual SLAM systems such as ORB-SLAM3~\cite{campos2021orbslam3} instead estimate pose and map incrementally in real time, but likewise triangulate over an extended trajectory rather than a single designated pair. EpiTransfer instead uses only two frames and
an externally supplied pose, avoiding both a fixed baseline and
iterative multi-frame optimization.

\subsection{Virtual and Temporal Stereo from Monocular Motion}

A separate line of work turns two frames of a single moving camera
into a stereo-like pair once camera motion is known, rather than
triangulating across many frames. This idea dates at least to
Nevatia's motion stereo \cite{nevatia1976depth}, which used closely
spaced monocular views along a vehicle's path in place of a physical
stereo baseline. In direct monocular SLAM, LSD-SLAM's temporal
stereo \cite{engel2014lsd} searches for pixel correspondences
along the epipolar line induced by the estimated relative pose
between the current and a reference keyframe, then triangulates
directly from the two \emph{real} camera centers -- the same
computation as our own Triangulation baseline (DLT) (Fig.~3), rather than
the virtual-view construction we propose. More recently,
deep-learning methods have revisited monocular-video-as-stereo
directly for depth: Brousseau and Roy \cite{brousseau2025spherical}
introduce a differentiable spherical epipolar rectification that
warps a monocular pair, under arbitrary known motion, into a
rectified configuration on which a self-supervised deep stereo
matcher computes disparity and absolute depth; VISTA
\cite{chengvista} similarly forms virtual stereo pairs -- in
their case from two vehicles' monocular views of the same scene --
to supply stereo-style supervision for training a monocular depth
network, rather than to produce depth directly at inference.

EpiTransfer differs from this line along two axes. First,
correspondence and view synthesis are sparse and feature-based
(XFeat) rather than dense: instead of a per-pixel rectifying warp of
the real images, we transfer sparse point correspondences into a
synthetic third view via closed-form trifocal intersection, so no
dense matching network or rectification stage is required. Second,
the virtual view's baseline is a free parameter decoupled from the
real inter-frame motion, which the dense-rectification methods above
do not exploit. This training-free, sparse formulation enables
real-time operation on embedded hardware, an explicit, geometrically derived characterization of
the method's failure mode near the trifocal plane -- a degeneracy
not characterized in the temporal- or virtual-stereo methods above.

\subsection{Learning-Based Monocular Depth for Aerial Imagery}
\label{sec:learning-based}
Learning-based monocular methods predict dense depth from a single
image or short video using visual priors, including self-supervised
approaches such as Monodepth2 \cite{godard2019diggingselfsupervisedmonoculardepth} and PackNet-SfM \cite{guizilini20203dpackingselfsupervisedmonocular} and general-purpose
metric-depth foundation models such as ZoeDepth
\cite{bhat2023zoedepthzeroshottransfercombining} and Depth Anything V2
\cite{yang2024depthv2}. Their reliability depends on the
match between deployment and training conditions, and UAV-specific
benchmarks show that altitude, viewing angle, and camera
characteristics induce substantial domain shift.

For oblique aerial imagery specifically, Madhuanand et al.
\cite{madhuanand2021self} train self-supervised depth and
pose networks from temporal UAV video using view reconstruction as
supervision, demonstrating depth prediction across large scene
depths but, like other learned methods, remaining tied to the
training distribution's scale and geometry. Geometric approaches
have also targeted this setting at shorter range: Chatzikalymnios
and Moustakas \cite{Chatzikalymnios2022} use downward-looking
physical stereo for landing-site detection, and Yang et al.
\cite{yang2018monocular} use ORB-SLAM triangulation with
barometer-scaled depth for autonomous landing, both evaluated up to
roughly 20--30\,m. EpiTransfer targets this regime at longer range (approximately 90~m, Section~V-A), using known pose rather than a physical baseline or barometric scale.

\subsection{Visual-Inertial Estimation and Aerial Navigation}
Visual-inertial odometry and SLAM systems such as VINS-Mono and ORB-SLAM3 estimate vehicle motion and maintain sparse feature maps for state estimation and map consistency, rather than per-feature range estimation for a designated image pair. EpiTransfer is complementary: it consumes an externally supplied pose estimate from such a system (or from GNSS/IMU) and focuses on the latter task.

% Along these lines, prior geometric methods require either a fixed
% physical baseline or costly multi-frame optimization, and prior
% virtual- and temporal-stereo methods construct depth from dense
% per-pixel matching -- either direct search along the real second
% view or a learned dense rectification-and-matching network.
% EpiTransfer instead combines sparse, learned feature correspondence
% with closed-form, training-free trifocal transfer, targeting the
% specific success and failure conditions of oblique aerial approach
% imagery.

\section{Method}
\label{sec:method}
\begin{figure}[t]
    \centering
    \includegraphics[width=\linewidth]{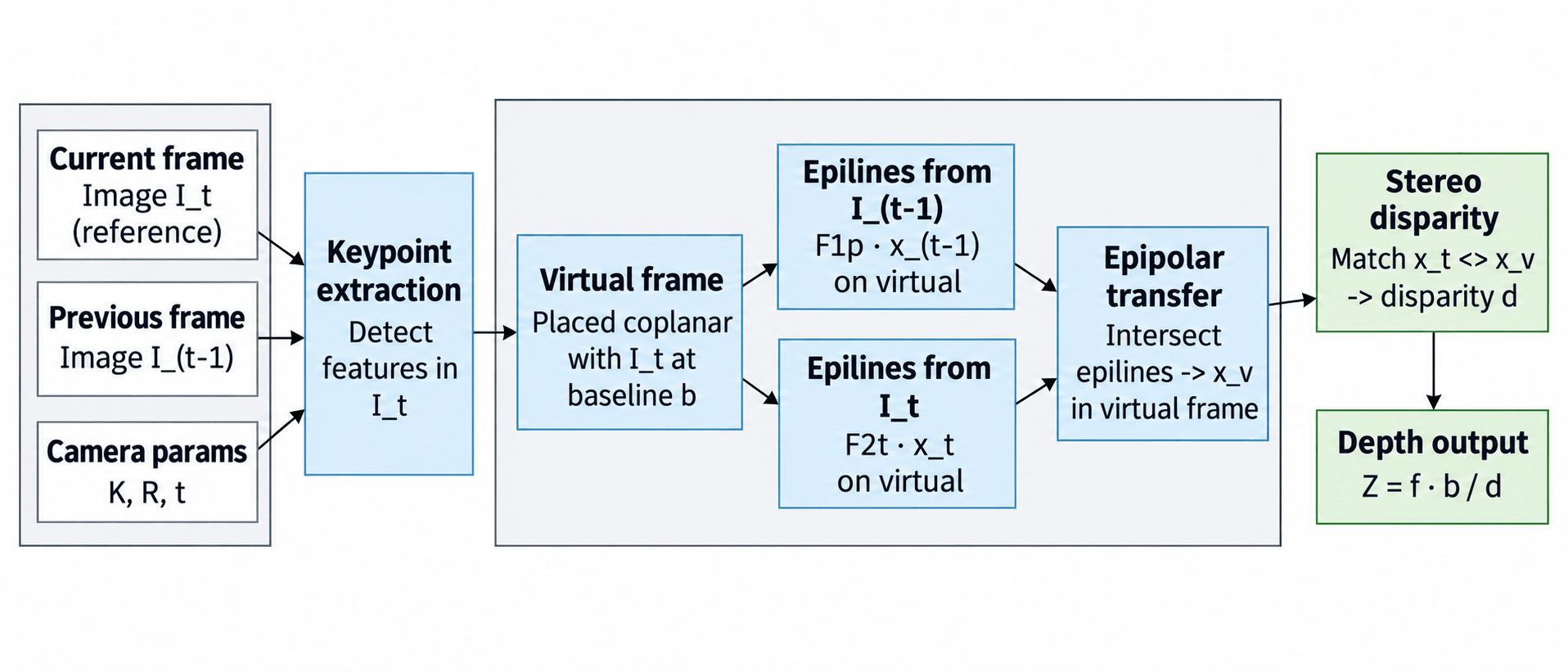}
    \caption{Flow chart of the proposed method, EpiTransfer}
    \label{fig:workflow}
\end{figure}

\begin{figure}[b]
    \centering
    \includegraphics[width=\linewidth]{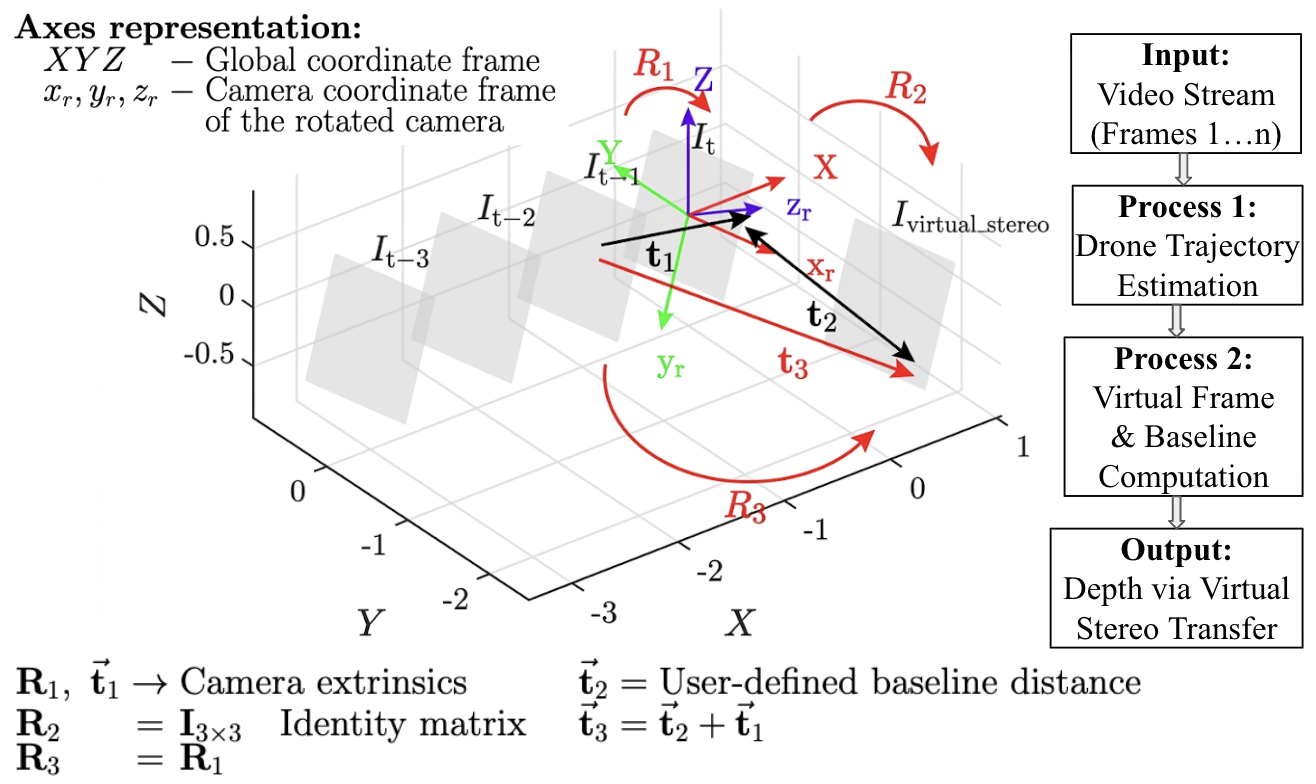}
    \caption{Virtual Stereo Frame synthesis using $I_t$ and $I_{t-1}$}
    \label{fig:synthesis}
\end{figure}

This section presents the proposed two-frame depth estimation framework. 
Given two temporally separated monocular images $I_t$ and $I_{t-1}$ with known relative pose, we synthesize a virtual stereo pair through epipolar transfer and recover depth using geometric triangulation. 
An overview of the pipeline is shown in Fig.~\ref{fig:workflow}, and virtual stereo frame synthesis is shown in Fig. ~\ref{fig:synthesis}.

\subsection{Problem Formulation}

Let $I_t$ and $I_{t-1}$ denote two RGB images captured by a calibrated monocular camera at different time instants. 
The relative camera motion between the frames is represented by rotation $\mathbf{R}$ and translation $\mathbf{t}$. 
The camera intrinsic matrix is denoted by $\mathbf{K}$.

Our objective is to estimate a dense depth map for $I_t$ using only $(I_t, I_{t-1})$ and the known camera motion.

\subsection{Feature Correspondence}

Reliable feature correspondences are essential to establish geometric constraints between views. Instead of traditional handcrafted algorithms like SIFT or ORB, we employ XFeat \cite{10655278} on $I_t$ and $I_{t-1}$. This yields robust sparse feature matches, even in low-texture regions, without requiring manual parameter tuning.

Since XFeat's sparse output rarely coincides with LiDAR-projected pixels, we instead use the dense RAFT feature matching algorithm~\cite{teed2020raftrecurrentallpairsfield} to obtain correspondences at every LiDAR-projected point for evaluation. Reported accuracy therefore reflects the triangulation geometry under dense correspondence, not the deployed sparse XFeat pipeline directly.

\subsection{Virtual Stereo frame using Epipolar transfer}

The key idea of our approach is to convert the two temporal views into a virtual stereo configuration.

Using the known intrinsics and relative pose, the fundamental matrix between two image frames can be computed as

\[
\mathbf{F} = \mathbf{K}^{-T} [\mathbf{t}]_\times \mathbf{R} \mathbf{K}^{-1},
\]

where $[\mathbf{t}]_\times$ denotes the skew-symmetric matrix of $\mathbf{t}$.

For each point $\mathbf{x}_t$ in $I_t$ and its correspondence $\mathbf{x}_{t-1}$ in $I_{t-1}$, the associated epipolar lines in a virtual view $V$ are given by

\[
\mathbf{l}_{t,V} = \mathbf{F}_{t,V} \mathbf{x}_{t}, \quad
\mathbf{l}_{t-1,V} = \mathbf{F}_{t-1,V} \mathbf{x}_{t-1} .
\]

where ${F}_{t,V}$ and ${F}_{t-1,V}$ are the fundamental matrix between $I_t$ and V, and $I_{t-1}$ and V, respectively.
The intersection of these epipolar lines determines the transferred location $\mathbf{x}_v$ in the virtual image:
\[
\mathbf{x}_v = \mathbf{l}_{t,V} \times \mathbf{l}_{t-1,V} .
\]
This operation effectively synthesizes a new viewpoint, forming a virtual stereo pair $(I_t, V)$ with a controllable baseline. 
Unlike physical stereo systems with a fixed baseline, the virtual baseline can be freely chosen according to the observed camera motion.

\subsection{Depth Triangulation}

Once the virtual correspondences $(\mathbf{x}_t, \mathbf{x}_v)$ are obtained, depth is recovered using standard stereo triangulation. 
Let $b$ denote the virtual baseline and $f$ the focal length. 
The depth $d$ of each pixel is computed from disparity $\Delta x$ as

\[
d = \frac{f b}{\Delta x}.
\]

This produces a depth map for $I_t$.

\subsection{Computational Complexity}

The proposed method requires only two frames and avoids bundle adjustment, global optimization, or training. The dominant cost arises from feature extraction and matching, enabling real-time performance on embedded robotic platforms. With XFeat features, the algorithm works at around 10 Hz on an Intel Core i7-12700H CPU.

\section{Experimental Setup}
\label{sec:setup}
This section describes the hardware platform, datasets, baselines, and evaluation metrics used to assess the proposed method.

\subsection{Hardware Platform}
Two experiments were conducted to evaluate the algorithm
across different operating scenarios. The first was performed
in an OptiTrack motion capture environment, in a room containing several objects of varying texture and structure. This
controlled experiment used a tripod-mounted rig equipped
with a Velodyne LiDAR and an Intel RealSense RGB-D
camera as shown in Fig. 4. This indoor environment serves as a controlled proxy for the oblique, non-forward camera motion characteristic of aerial approach trajectories.
% Two experiments were conducted to evaluate the algorithm across different operating scenarios. The first was performed in an OptiTrack motion capture environment, in a room containing several objects of varying texture and structure. This controlled experiment used a tripod-mounted rig equipped with a Velodyne LiDAR and an Intel RealSense RGB-D camera as shown in Fig. \ref{fig:indoor-rig}.

\begin{figure}
    \centering
    \includegraphics[width=0.4\linewidth]{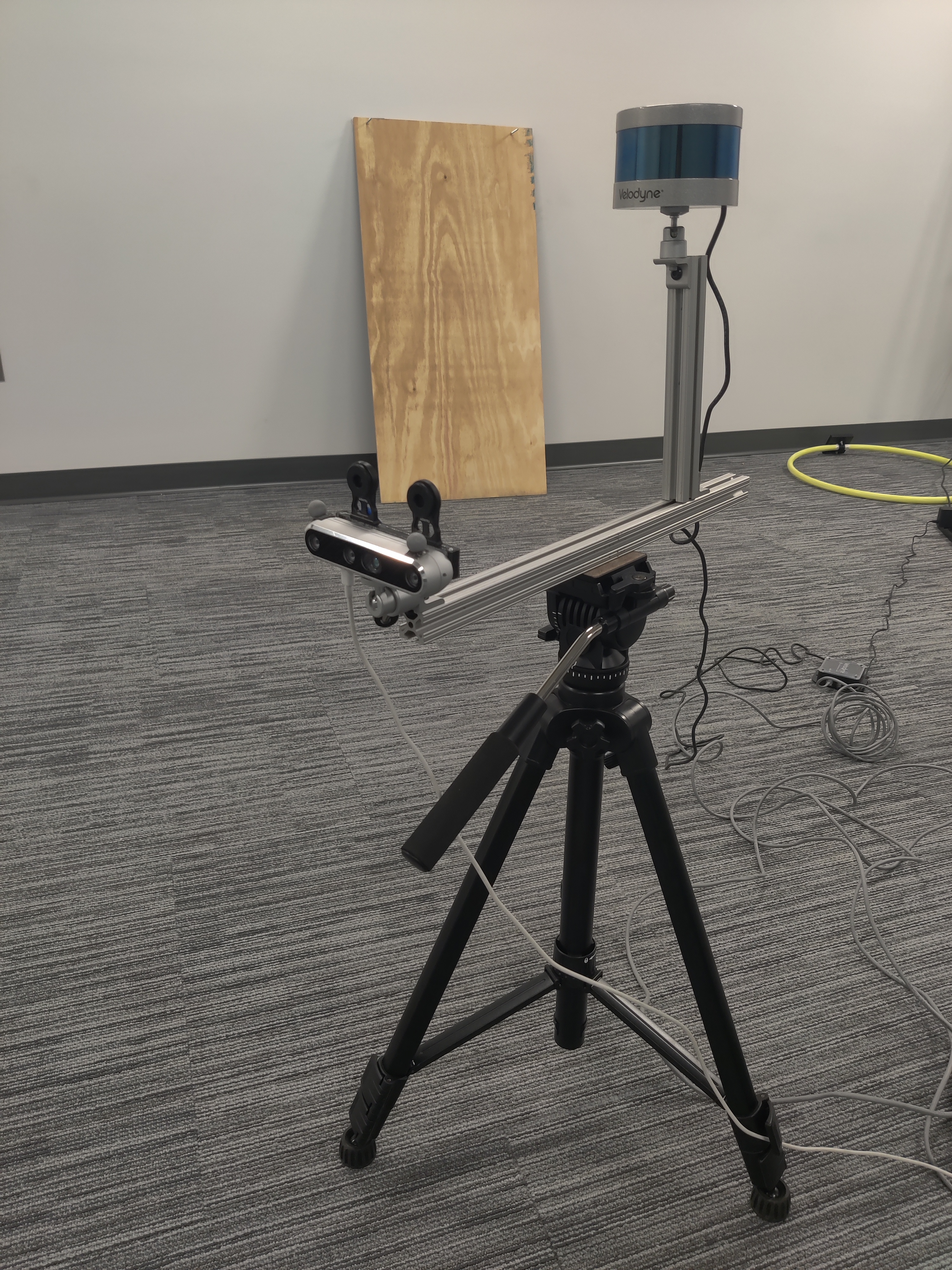}
    \caption{Experiment setup with camera and LiDAR mounted on the tripod in the indoor OptiTrack motion capture environment  }
    \label{fig:indoor-rig}
\end{figure}
The second experiment was conducted outdoors at an outdoor test facility using a DJI Mavic Air 2 \cite{dji} drone equipped with an RGB camera. Images were collected at 5-meter intervals and processed offline for the results reported in this paper. The camera was mounted at a fixed 20\textdegree{} downward pitch, with yaw matched to the drone's heading.

In addition to the offline outdoor evaluation, we benchmarked
real-time onboard performance of the pipeline on an NVIDIA AGX
Orin, the compute platform for our hexacopter test bed, with the
camera configured for the gimbal-stabilized 20\textdegree{}
downward pitch used in our outdoor flights. Using live camera
input on the ground, the full pipeline runs onboard at 10\,Hz on
this hardware. Flight testing on this platform is in progress at
the time of writing; a demonstration video will be included as
supplementary material. 
\subsection{Data Collection}
\label{sec:data_collection}

We collected indoor and outdoor sequences containing forward motion, rotations, and varying scene structure, including low-texture regions. The camera was operated in monocular mode, and only two consecutive RGB frames were used for depth estimation at each timestep.

For the outdoor dataset, 8 images were used for quantitative evaluation, of which 5 are shown in Fig.~\ref{fig:qualitative_depth_comparison_dji} for qualitative comparison. The camera pitch was held constant at $20\deg$ below the horizon throughout data collection.

For the indoor dataset, 7 images were used for evaluation. Two camera orientations were tested: forward-facing (pitch level with the horizon) and downward-pitched ($30\deg$ below the horizon). In both cases, camera motion between frames included lateral and rotational components rather than pure forward translation, ensuring the epipole remained outside the image region and avoiding the degenerate configuration discussed.

In both the indoor and outdoor datasets, image pairs were selected to span a range of approach angles and camera rotations.

\subsection{Pose Estimation}

Our method requires the relative position and orientation of the camera between consecutive frames. For the outdoor experiment, camera pose was obtained from the Pixhawk flight controller's onboard GPS and IMU. For the indoor experiment, camera pose was obtained from the OptiTrack motion capture system.

% Since our method's depth accuracy is directly coupled to the accuracy of the estimated inter-frame translation (Section~\ref{sec:method}), we characterize pose accuracy for both platforms. The OptiTrack system provides position accuracy on the order of $\pm$0.15 mm to $\pm$0.2 mm and orientation accuracy on the order of XX\textdegree{} at XX Hz, which we treat as ground truth indoors. The Pixhawk GPS+IMU provides position accuracy of approximately XX.XX m and orientation accuracy of approximately XX.XX\textdegree{}. At a typical flight speed of XX m/s and 10 Hz frame rate, this corresponds to an inter-frame baseline of approximately XX.XX m, giving a translation uncertainty of roughly XX\% of the baseline for the outdoor experiment.
\subsection{Comparison}

We compare against two baseline classes. First, off-the-shelf
learning-based monocular depth models, ZoeDepth
\cite{bhat2023zoedepthzeroshottransfercombining} and Depth Anything V2
\cite{yang2024depthv2}, which produce metric depth from a single RGB
image without pose and are used here without any fine-tuning on
aerial or oblique imagery, characterizing the out-of-the-box
domain-generalization gap of pretrained models rather than a
best-case baseline and highlighting that EpiTransfer requires no
training data. Second, direct two-view triangulation (DLT)
\cite{dlt}, which consumes the same correspondences and pose as
EpiTransfer but triangulates directly from the two
real camera centers rather than a synthetic virtual view, isolating the effect of our epipolar-transfer
formulation from the effect of having accurate pose.
\subsection{Evaluation Metrics}
We report standard depth metrics: AbsRel, SqRel, RMSE, RMSE$_{\log}$, MAE, and threshold accuracy $\delta < 1.25^k$ for $k=1,2,3$, computed identically across methods on matched valid-pixel masks.

\subsection{Ground Truth}
\textbf{Outdoor}. We evaluate outdoor depth estimates against the USGS 2017 LiDAR survey~\cite{USGS}. The georeferenced LiDAR cloud was converted to a local NED frame centered at the camera and projected into each image using the GPS/IMU-derived camera pose, gimbal attitude, and calibrated intrinsics. Since the LiDAR cloud and image trajectory are independently georeferenced, we did not use ICP; instead, a single constant translation, estimated by least squares from four manually identified control points, corrected a residual global position bias, reducing control-point reprojection RMS from 463~px to 32~px (control points were excluded from evaluation). Projected points were visibility-filtered via a z-buffer retaining the nearest LiDAR return per bin, with isolated outliers and edge-adjacent points removed. Depth was evaluated up to a maximum ground-truth range of 100\,m;
beyond this range, and in regions without a valid LiDAR return, no
depth estimate was scored for any method. The remaining visible LiDAR projections define the outdoor reference depths.

\textbf{Indoor.}
Indoor reference depths were obtained from a Velodyne VLP-16 Puck \cite{lidar} rigidly mounted with the RealSense camera on a tripod. The LiDAR-camera extrinsic transform was estimated by planar target-based calibration and achieved a mean reprojection error of 3 px \cite{verma2019automaticextrinsiccalibrationcamera}. LiDAR points were transformed to the camera frame using the synchronized OptiTrack pose of the sensor rig and then projected into the camera image.

For both datasets, evaluation was restricted to projected LiDAR returns within the shared LiDAR--camera field of view and the specified valid-depth range.

\section{RESULTS}
\label{sec:results}
\subsection{Outdoor}
\label{sec:res_outdoor}
\begin{figure*}[t]
\centering
\small
% Column Header Titles
\begin{tabular}{ccccc}
    \makebox[0.18\textwidth]{\textbf{Input Image}} &
    \makebox[0.18\textwidth]{\textbf{EpiTransfer (Ours)}} &
    \makebox[0.18\textwidth]{\textbf{Triangulation}} &
    \makebox[0.18\textwidth]{\textbf{ZoeDepth}} &
    \makebox[0.17\textwidth]{\textbf{Depth Anything}} \\
\end{tabular}

\vspace{1pt}

% ---------------- Row 1 ----------------
\includegraphics[width=0.19\textwidth]{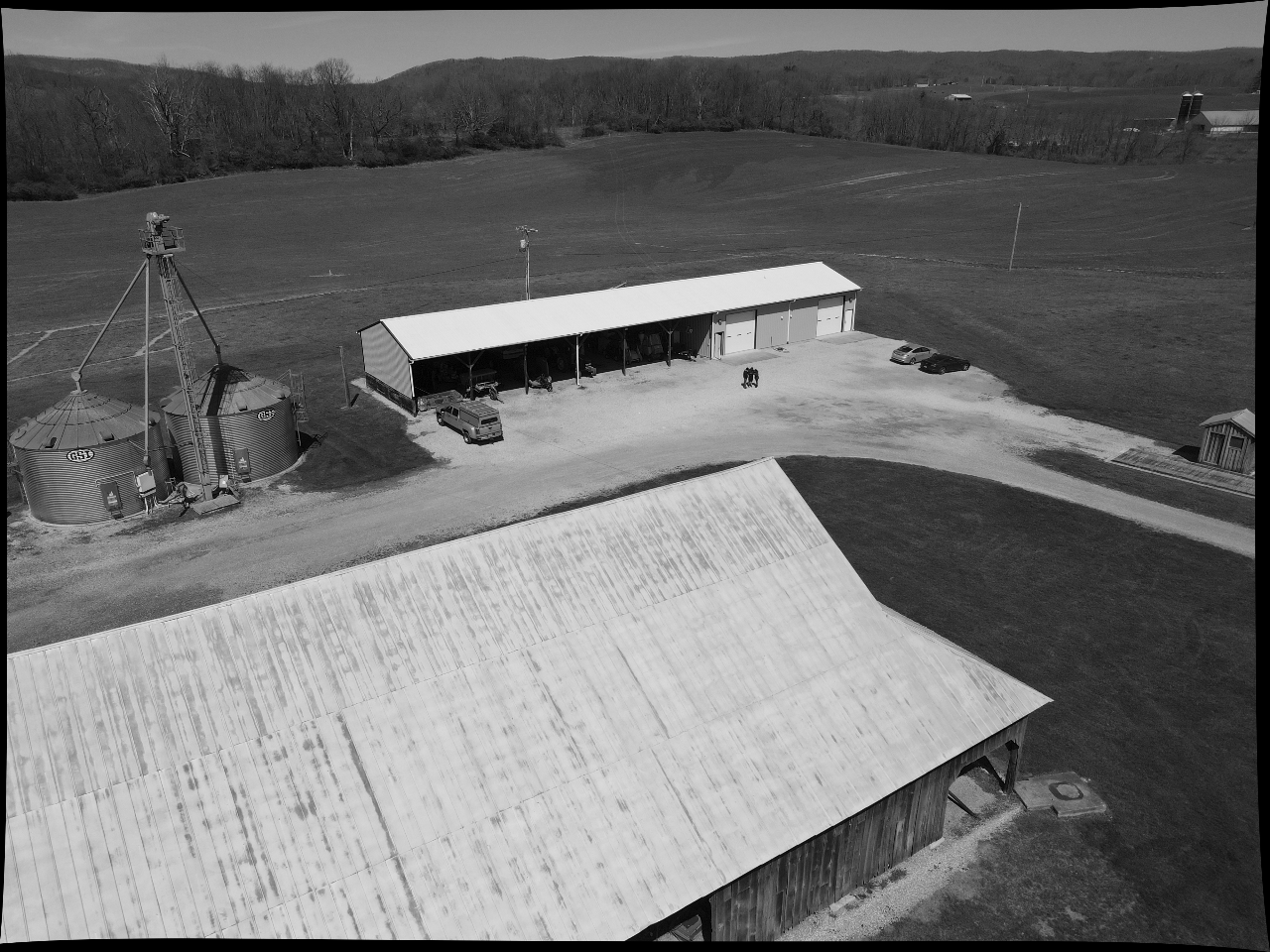}\hfill
\includegraphics[width=0.19\textwidth]{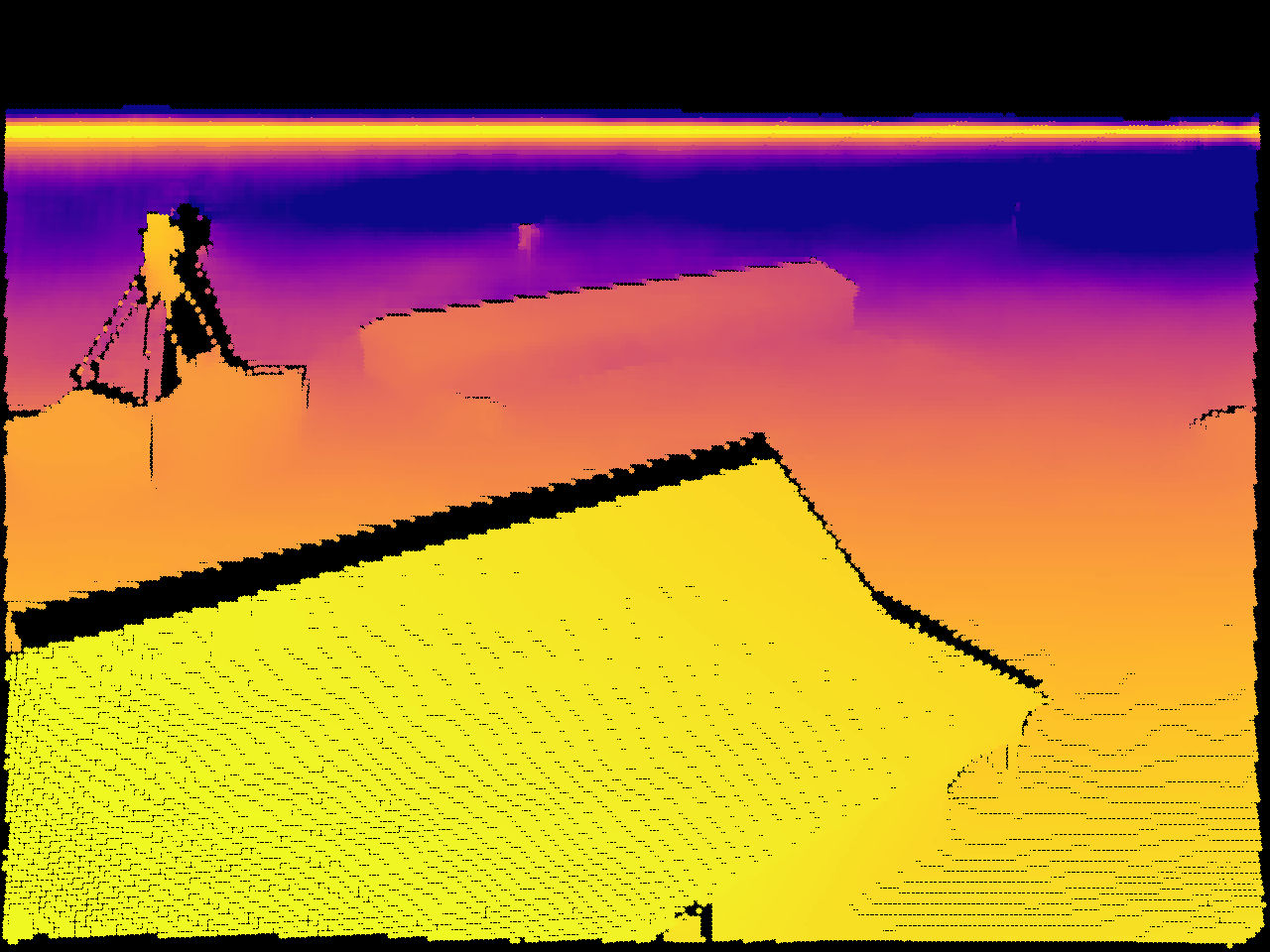}\hfill
\includegraphics[width=0.19\textwidth]{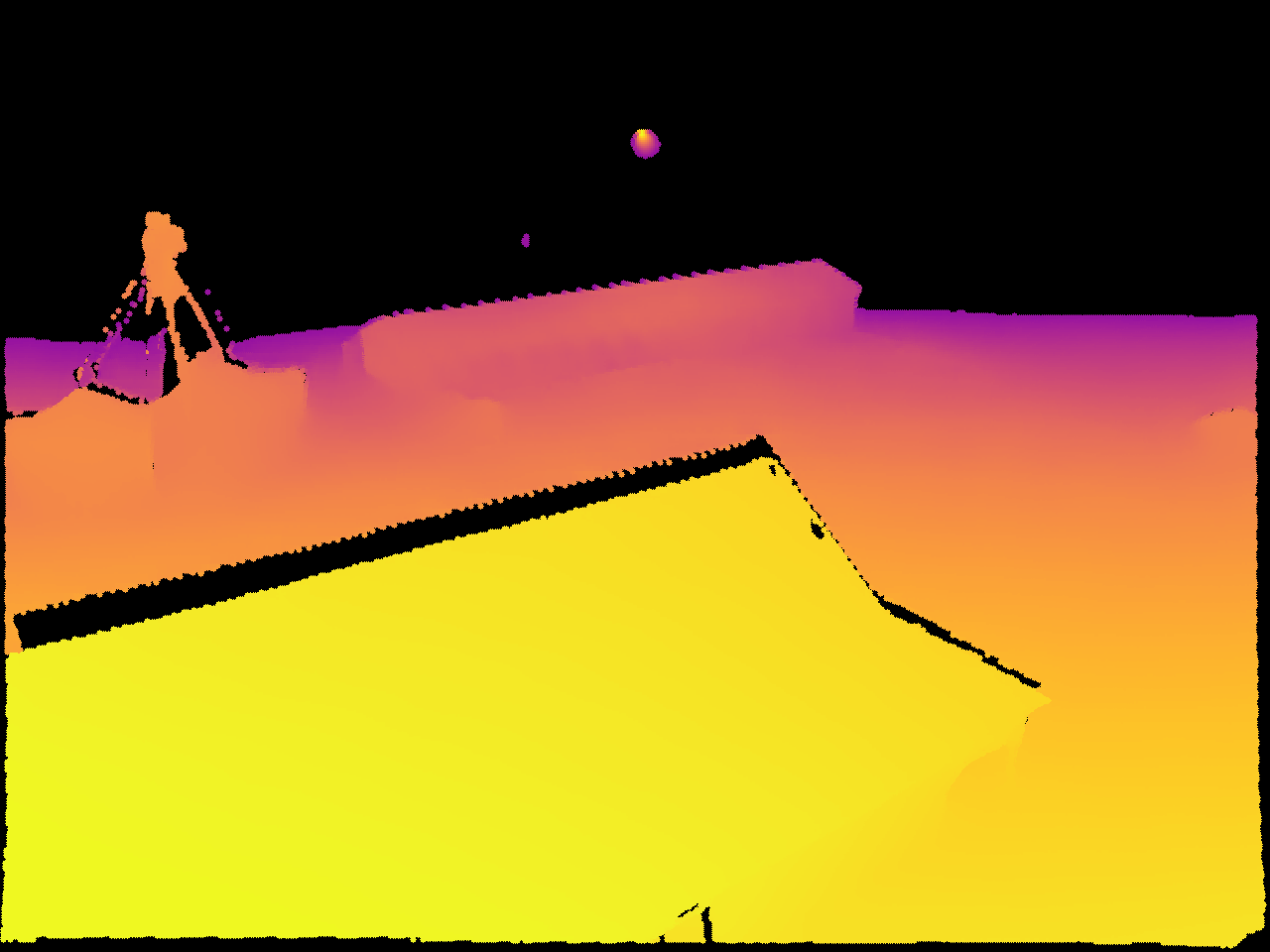}\hfill
\includegraphics[width=0.19\textwidth]{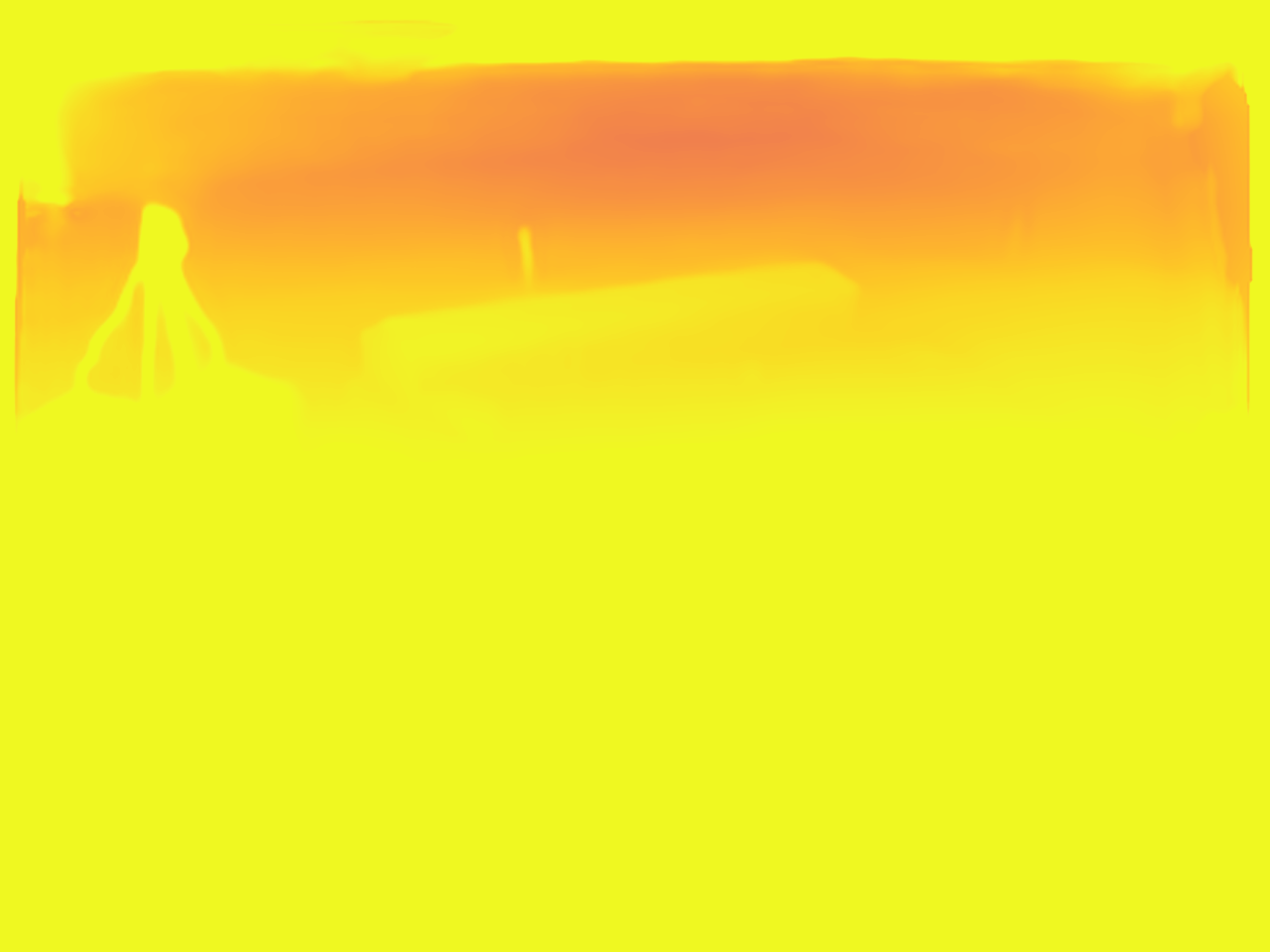}\hfill
\includegraphics[width=0.19\textwidth]{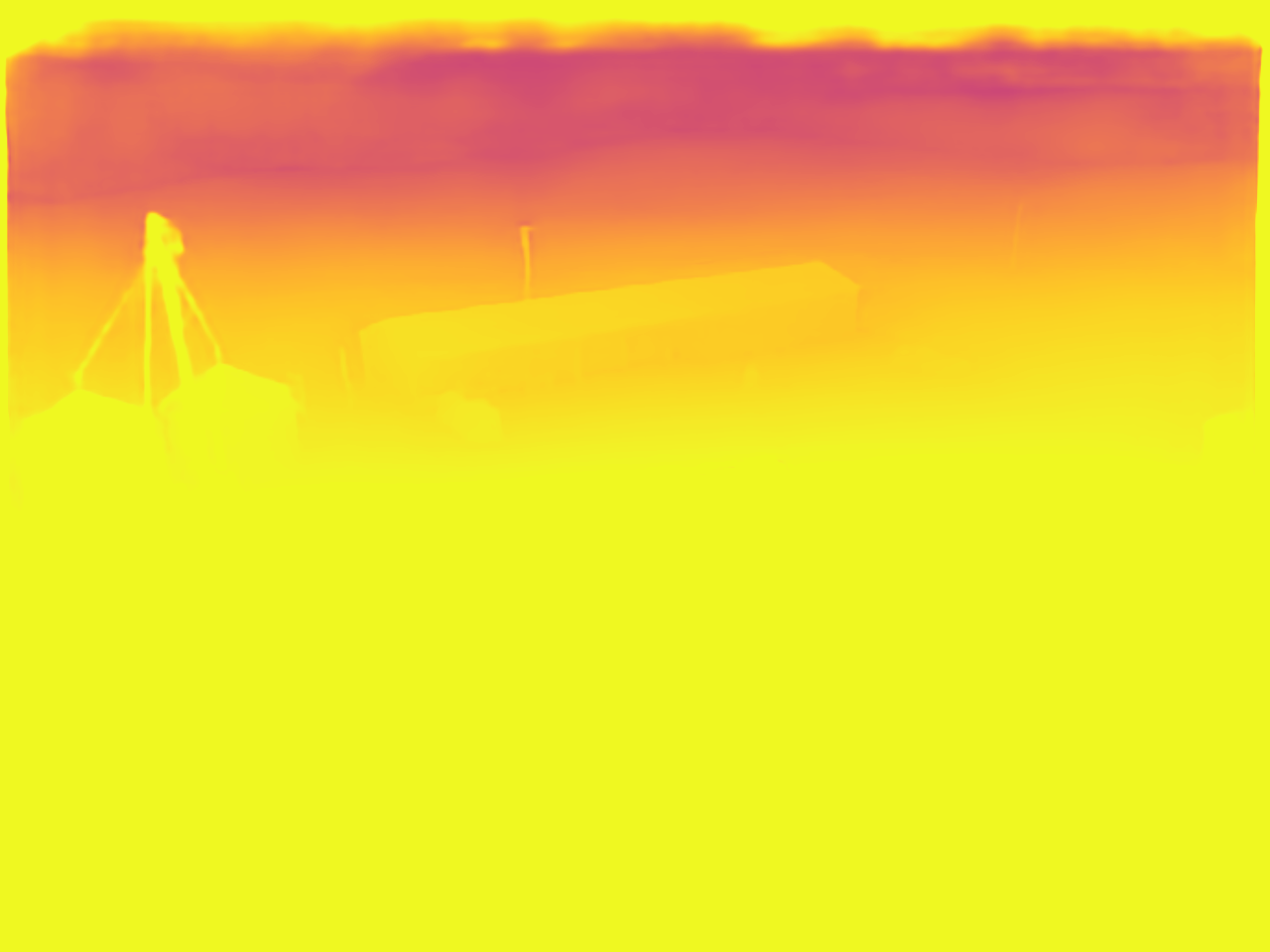}

\vspace{1pt}

% % ---------------- Row 2 ----------------
\includegraphics[width=0.19\textwidth]{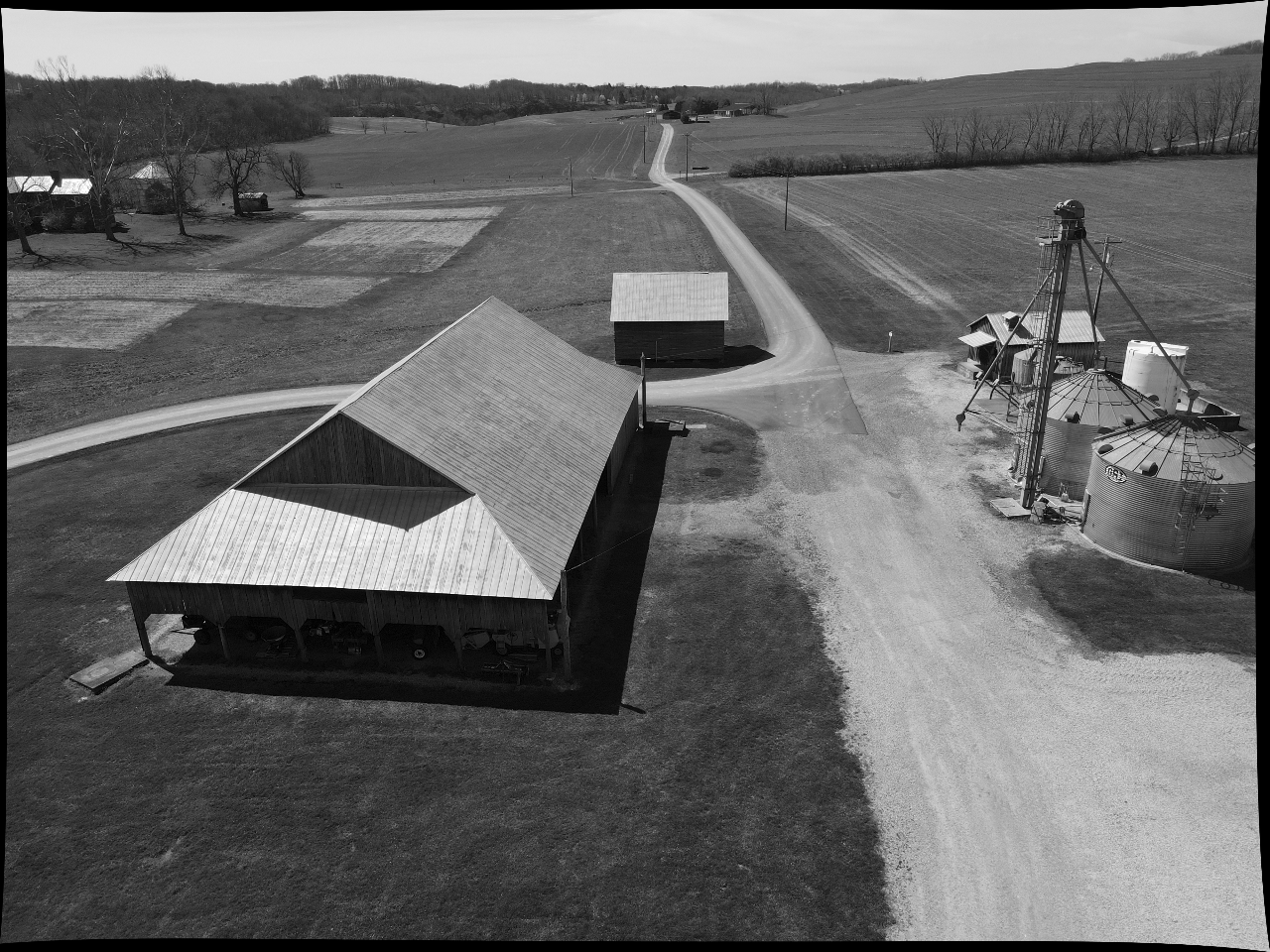}\hfill
\includegraphics[width=0.19\textwidth]{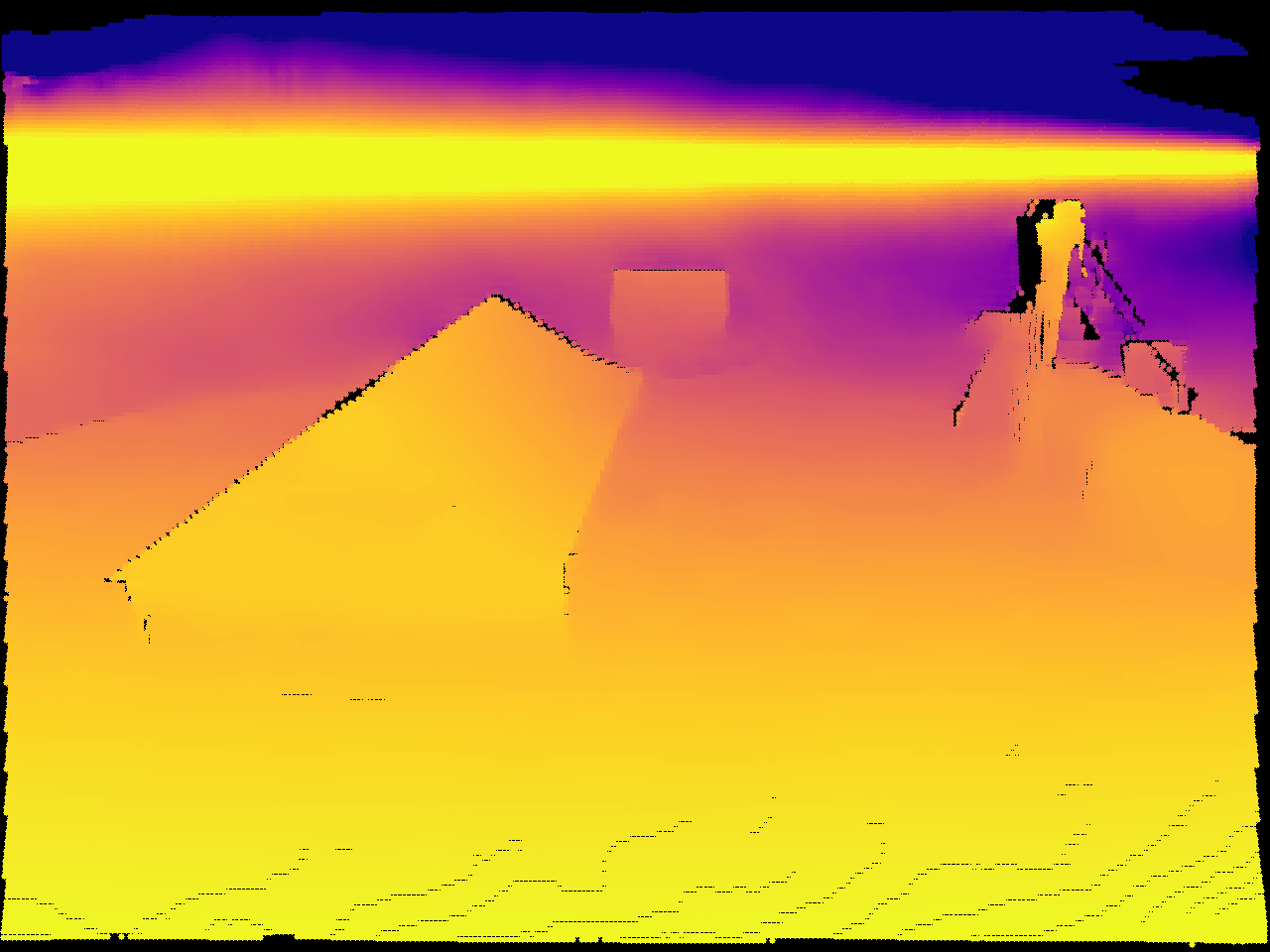}\hfill
\includegraphics[width=0.19\textwidth]{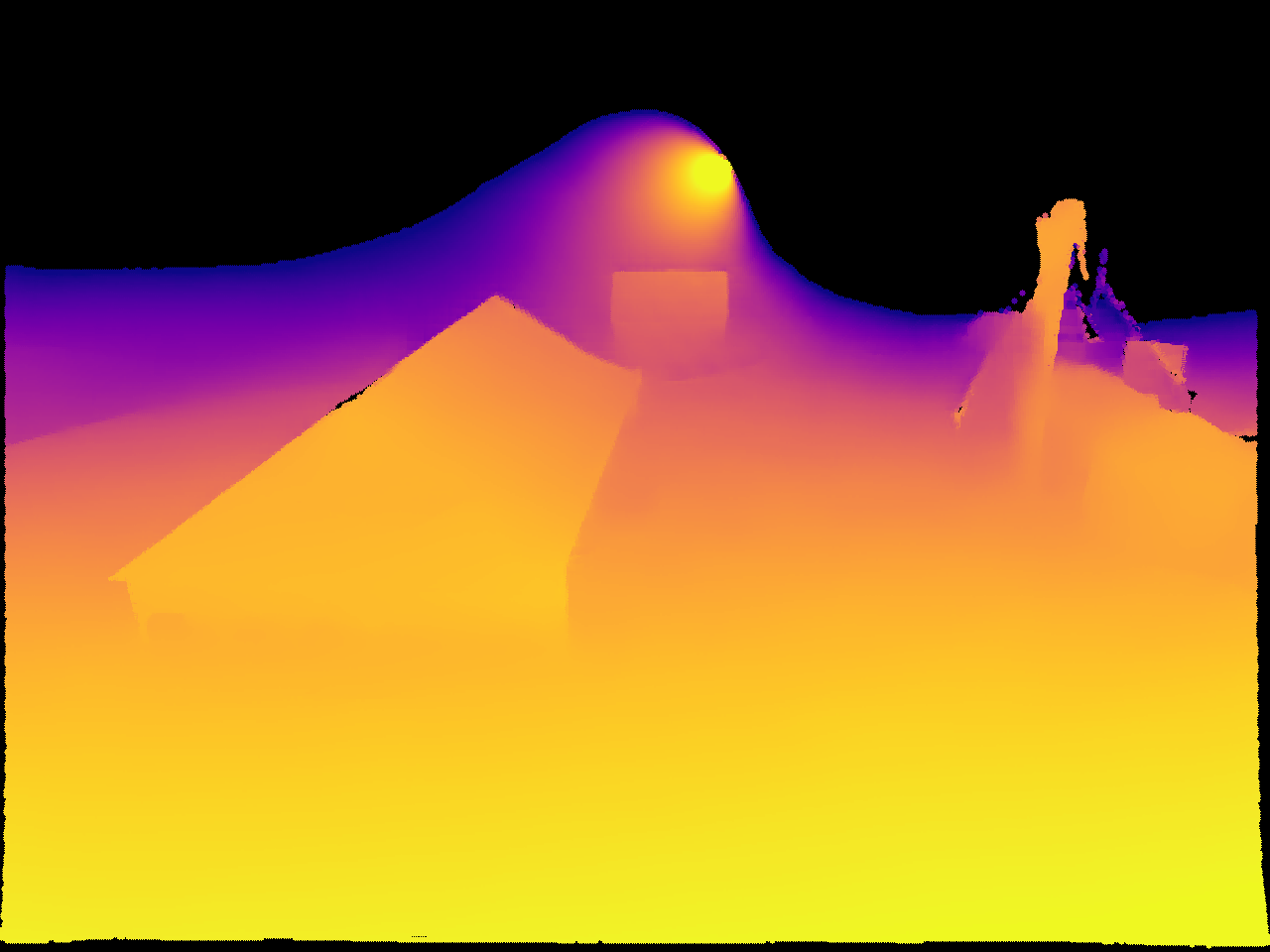}\hfill
\includegraphics[width=0.19\textwidth]{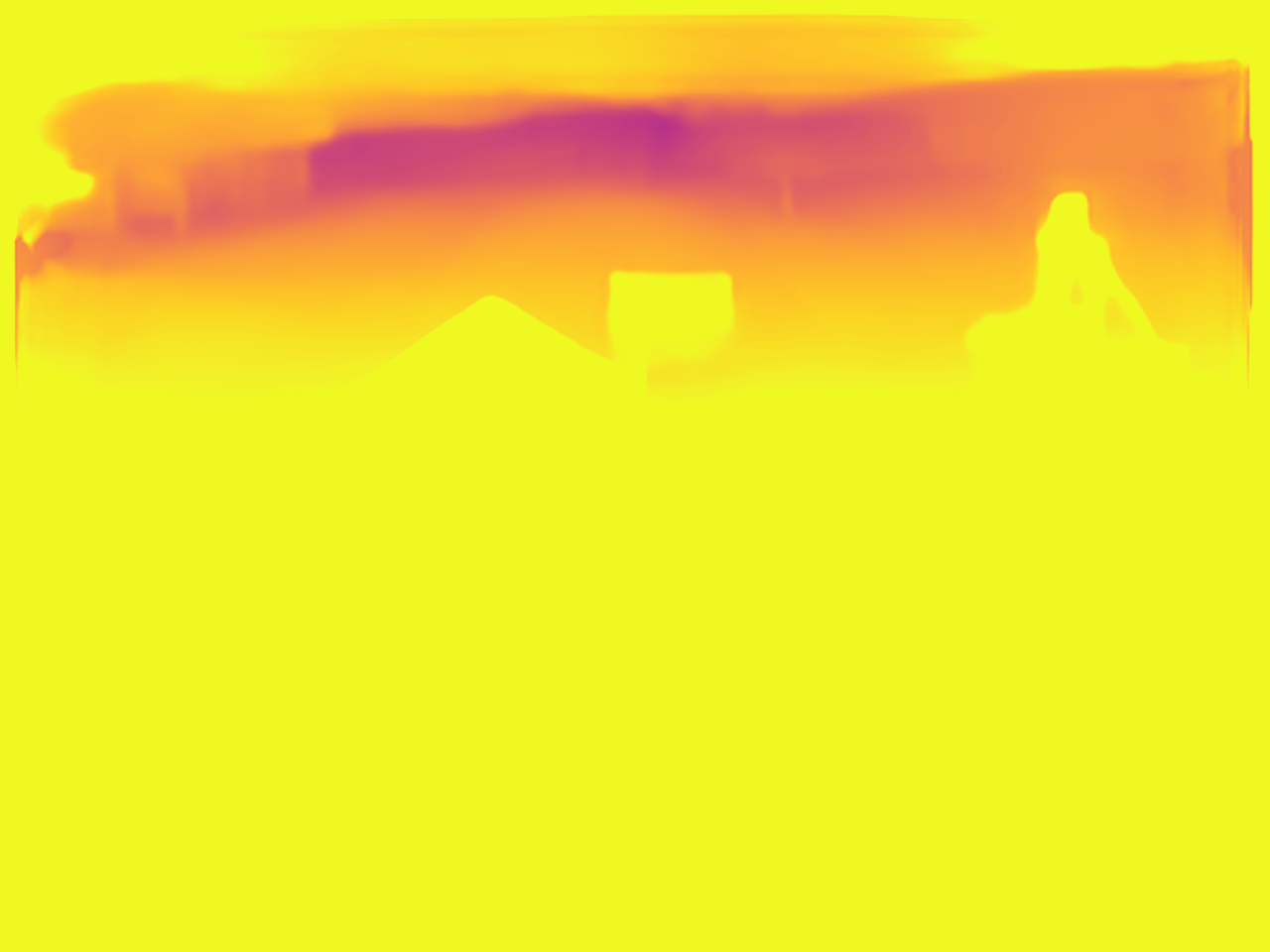}\hfill
\includegraphics[width=0.19\textwidth]{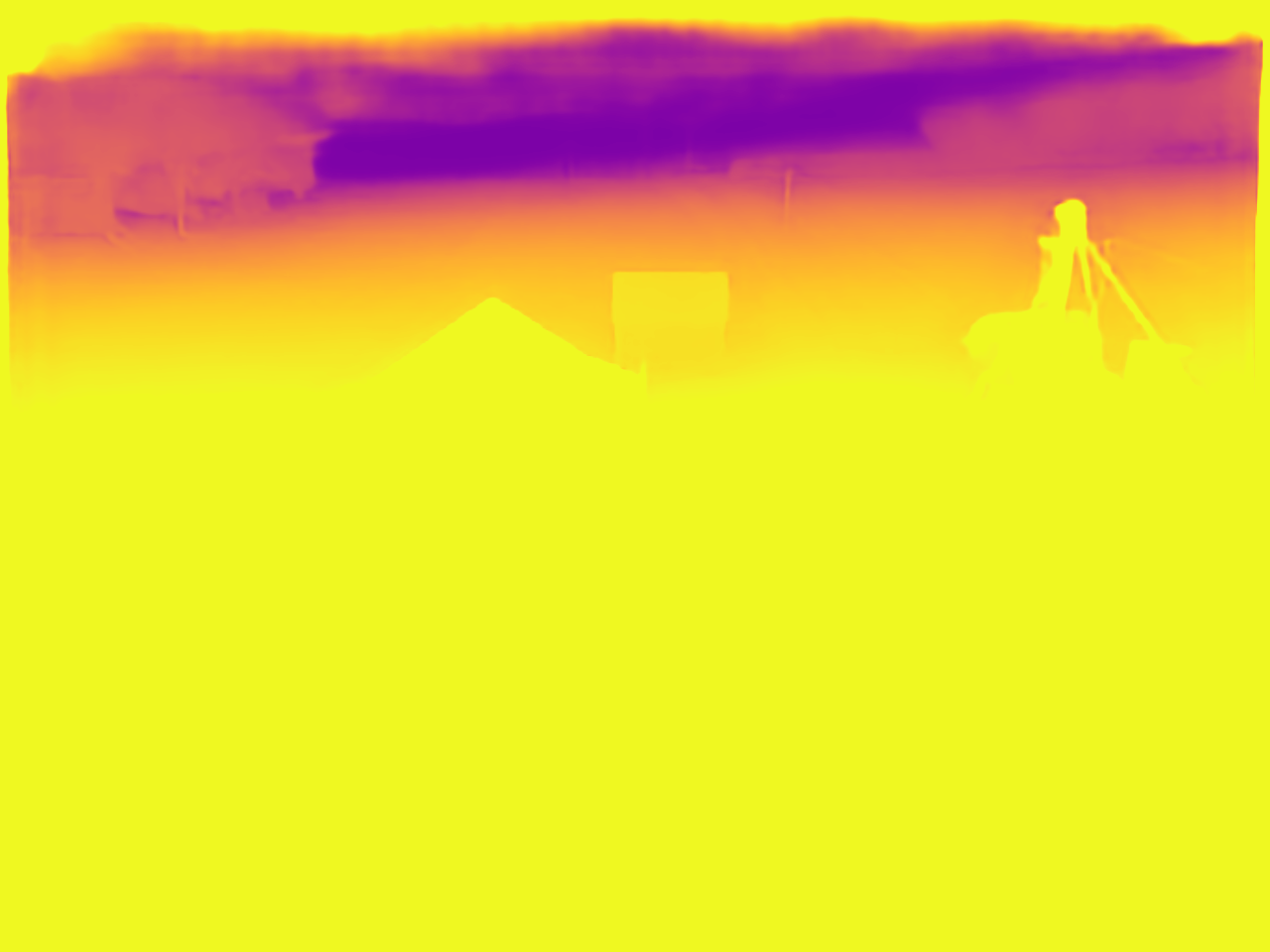}

\vspace{1pt}

% % ---------------- Row 3 ----------------
\includegraphics[width=0.19\textwidth]{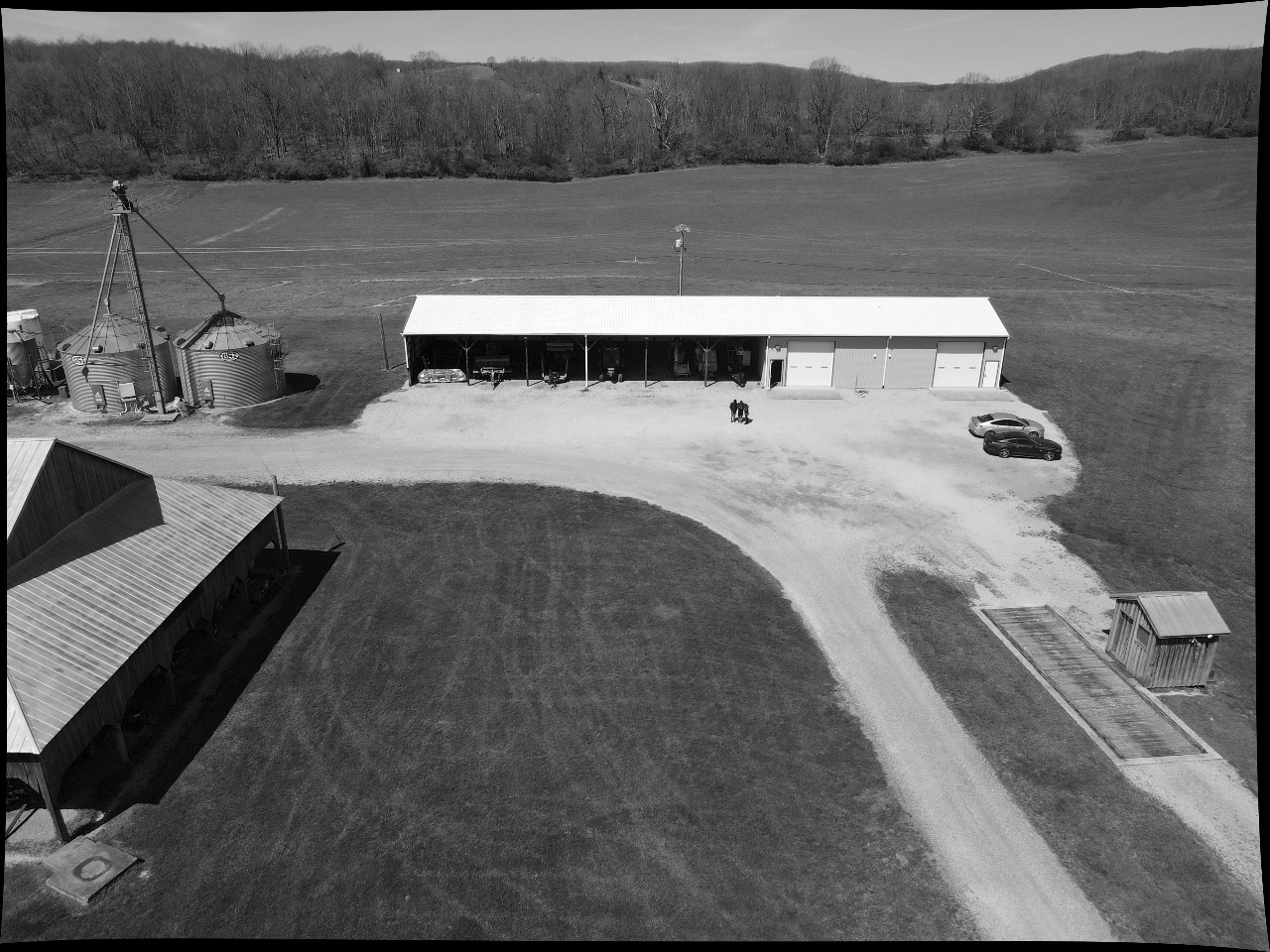}\hfill
\includegraphics[width=0.19\textwidth]{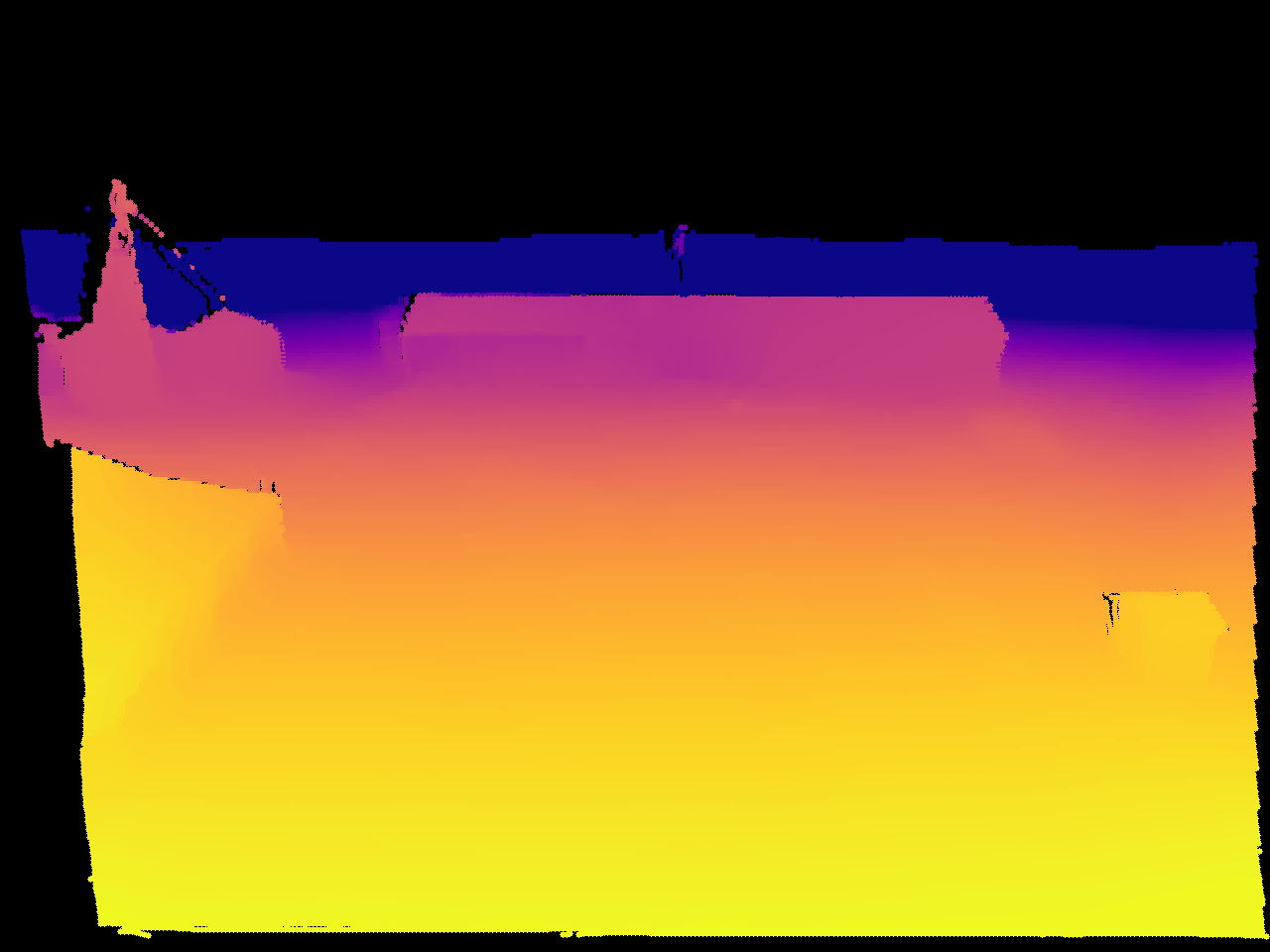}\hfill
\includegraphics[width=0.19\textwidth]{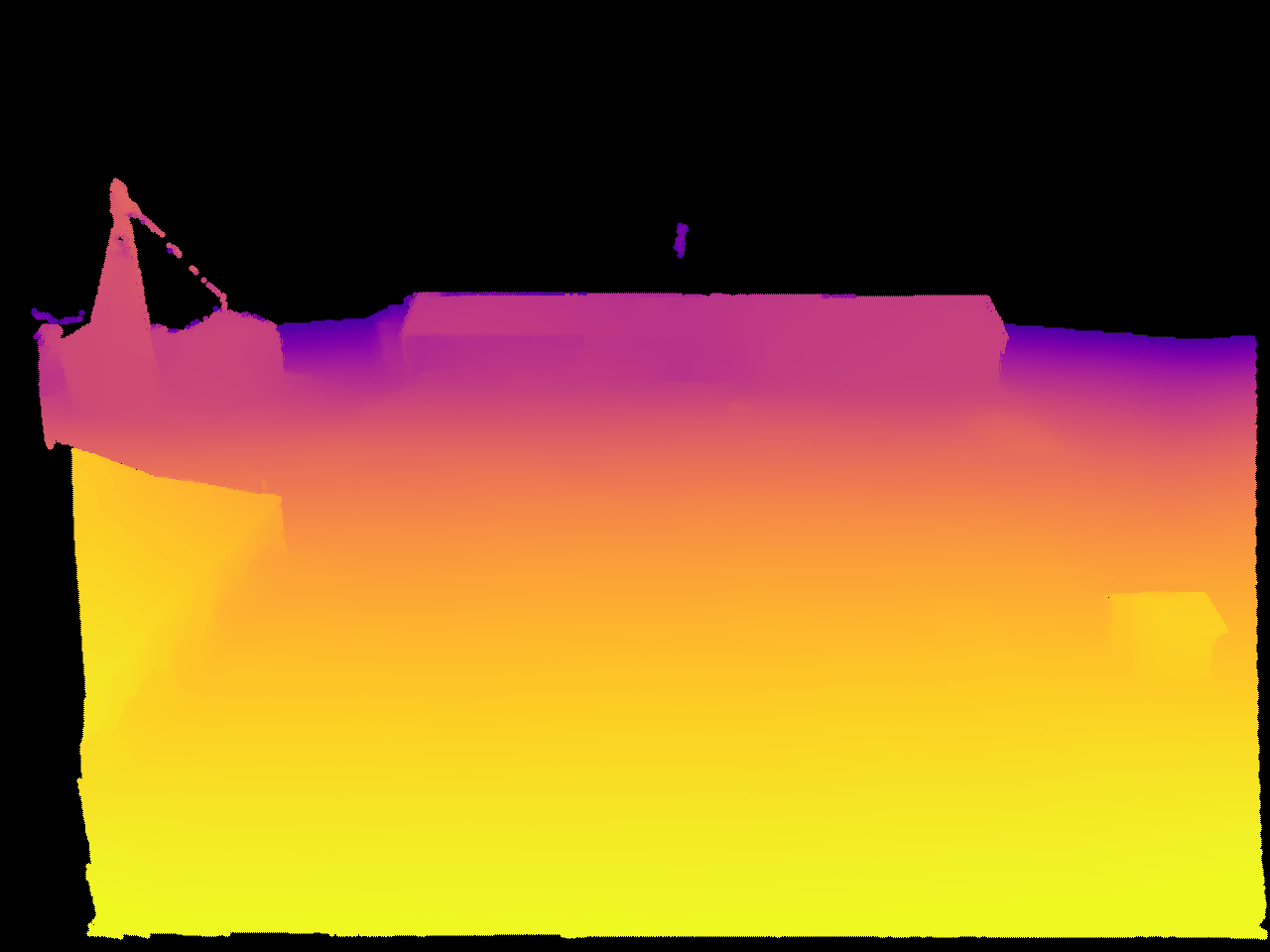}\hfill
\includegraphics[width=0.19\textwidth]{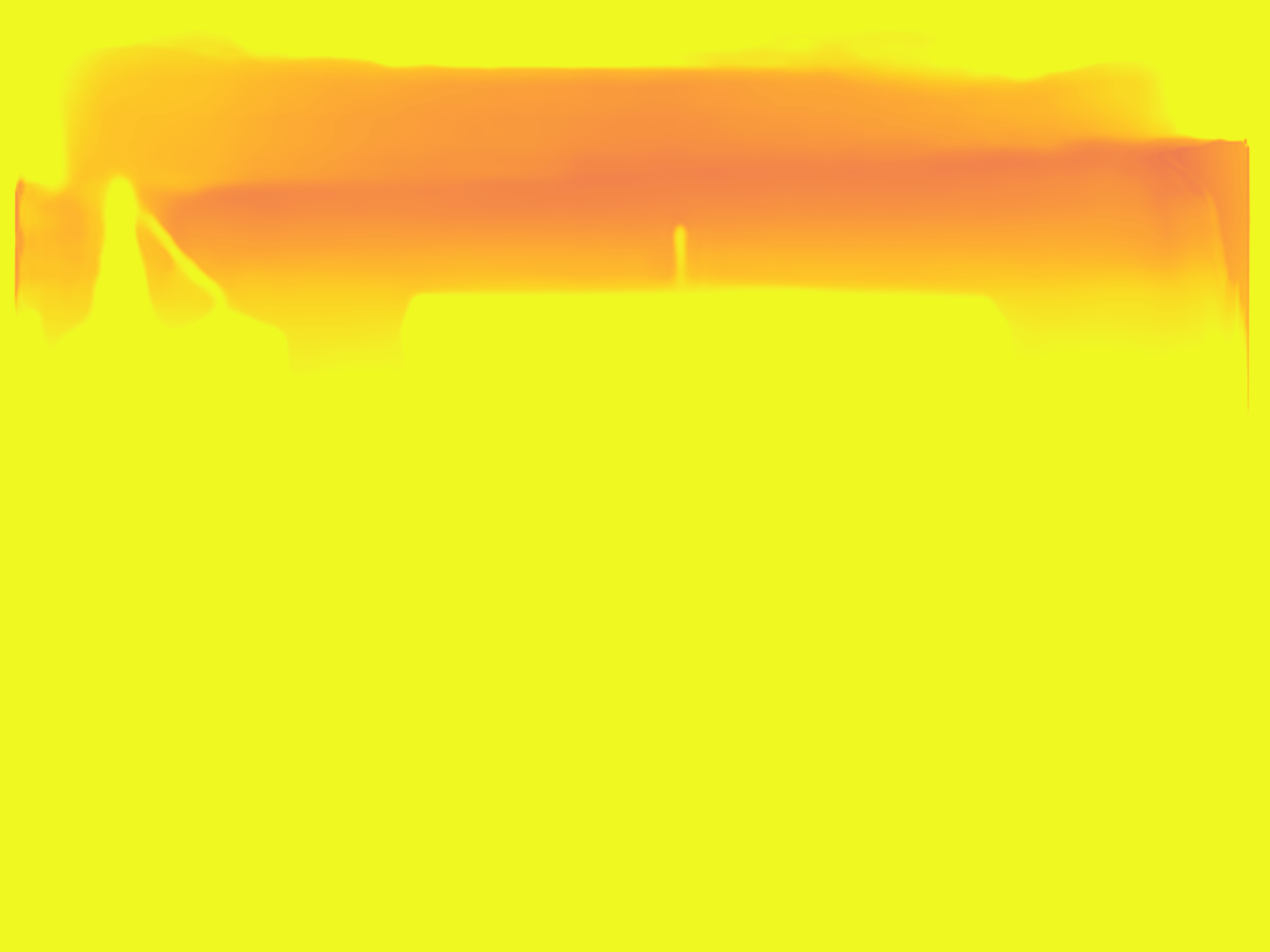}\hfill
\includegraphics[width=0.19\textwidth]{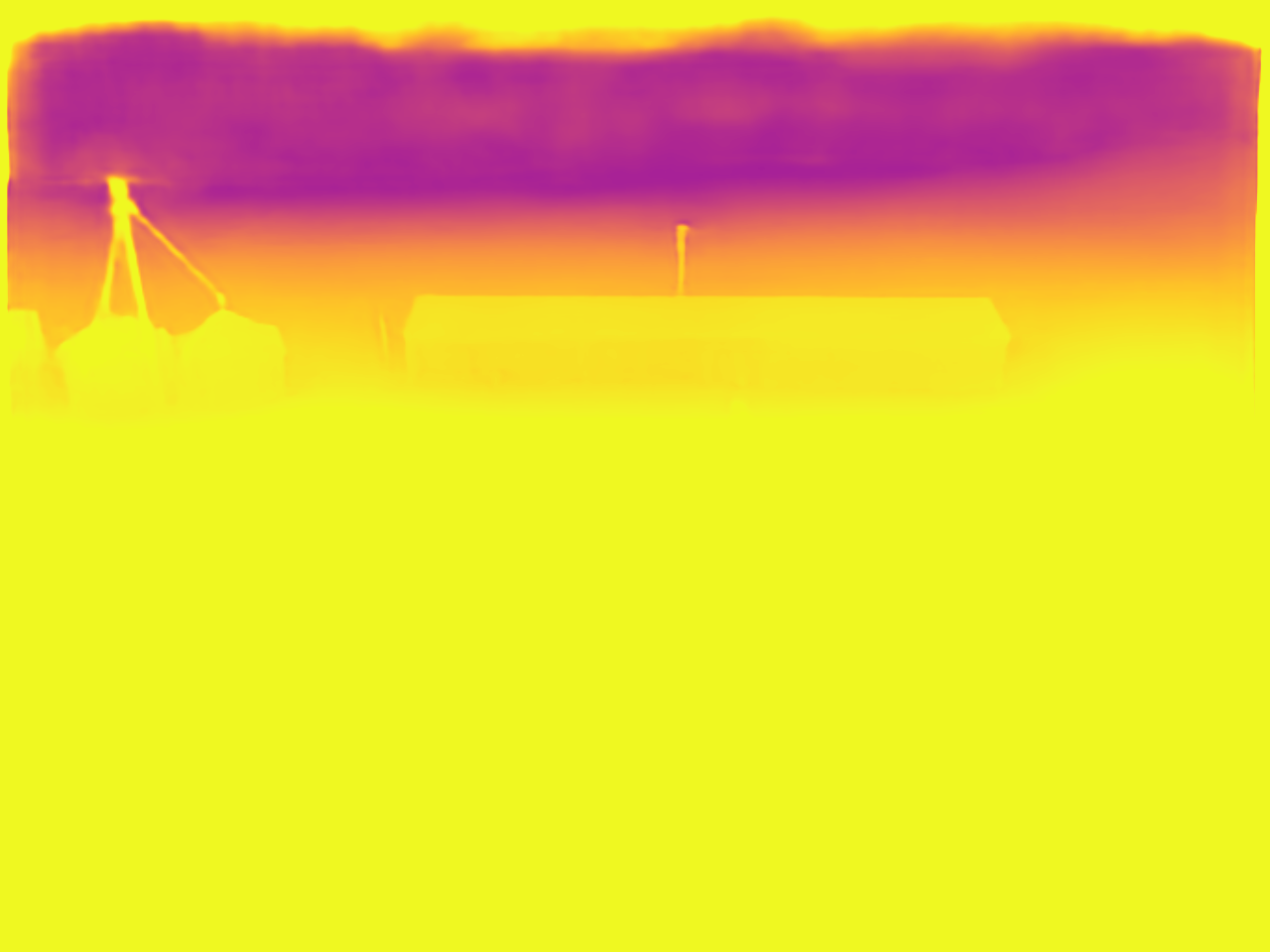}
\vspace{1pt}

% % ---------------- Row 3 ----------------

\includegraphics[width=0.19\textwidth]{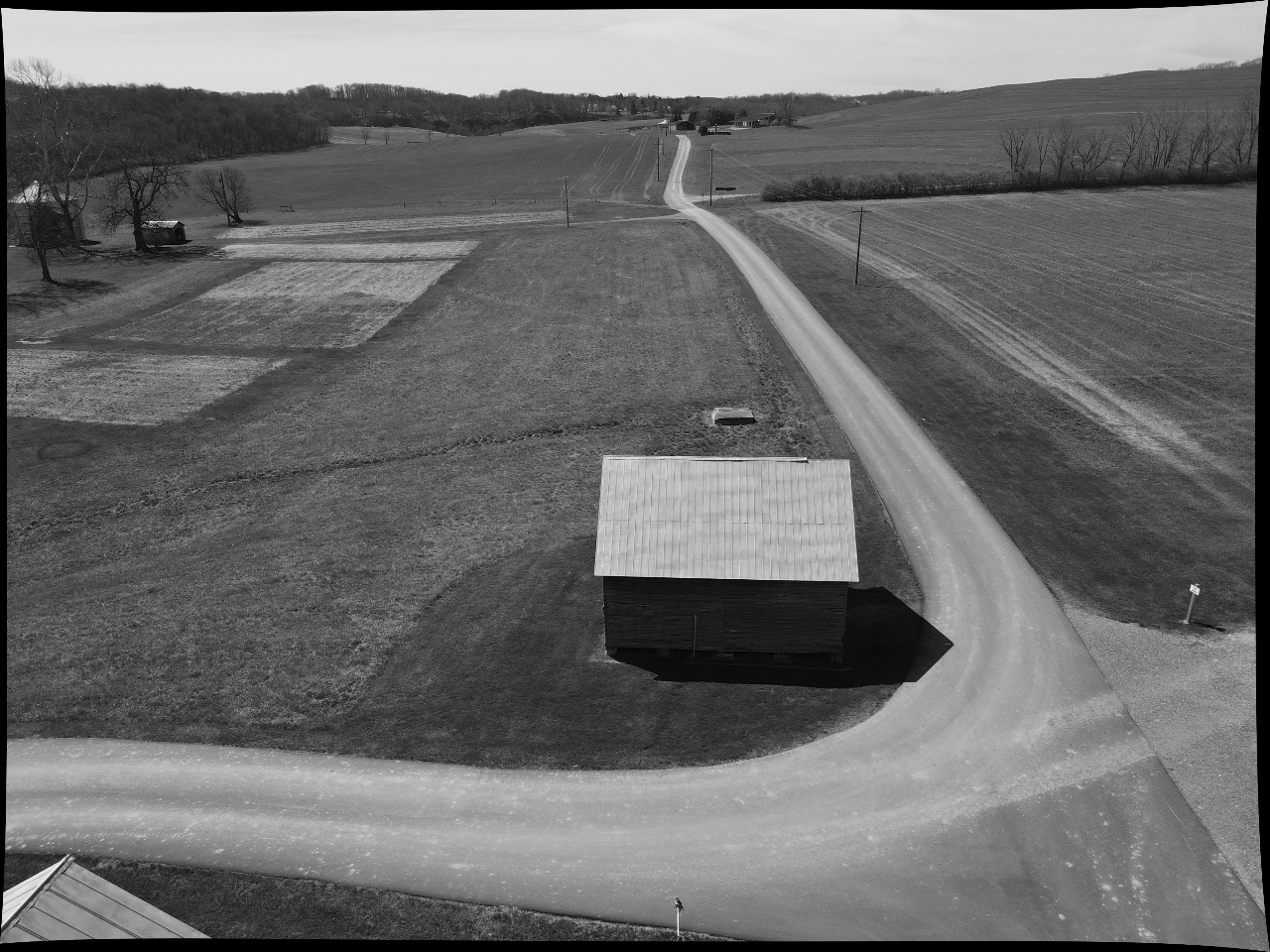}\hfill
\includegraphics[width=0.19\textwidth]{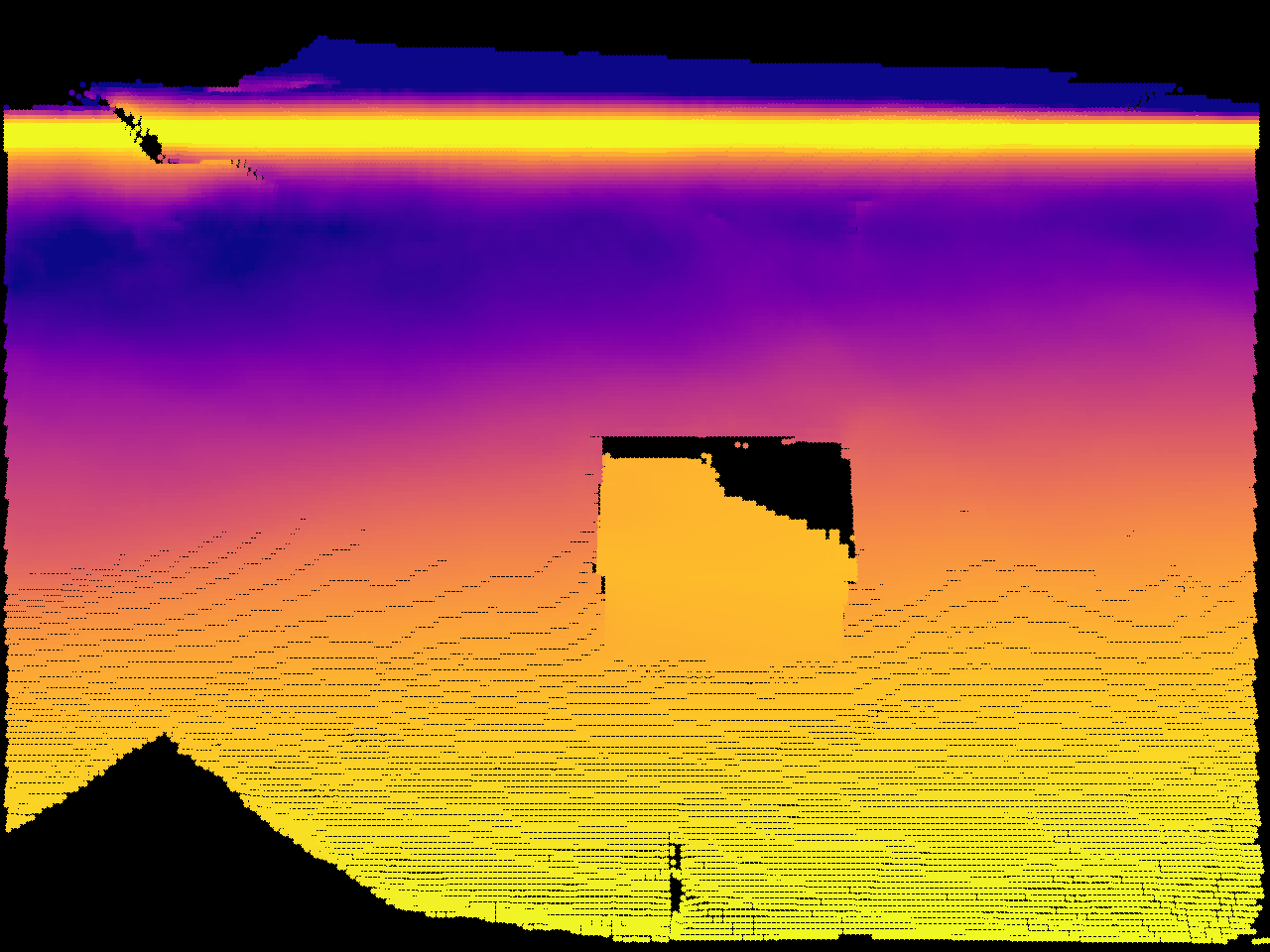}\hfill
\includegraphics[width=0.19\textwidth]{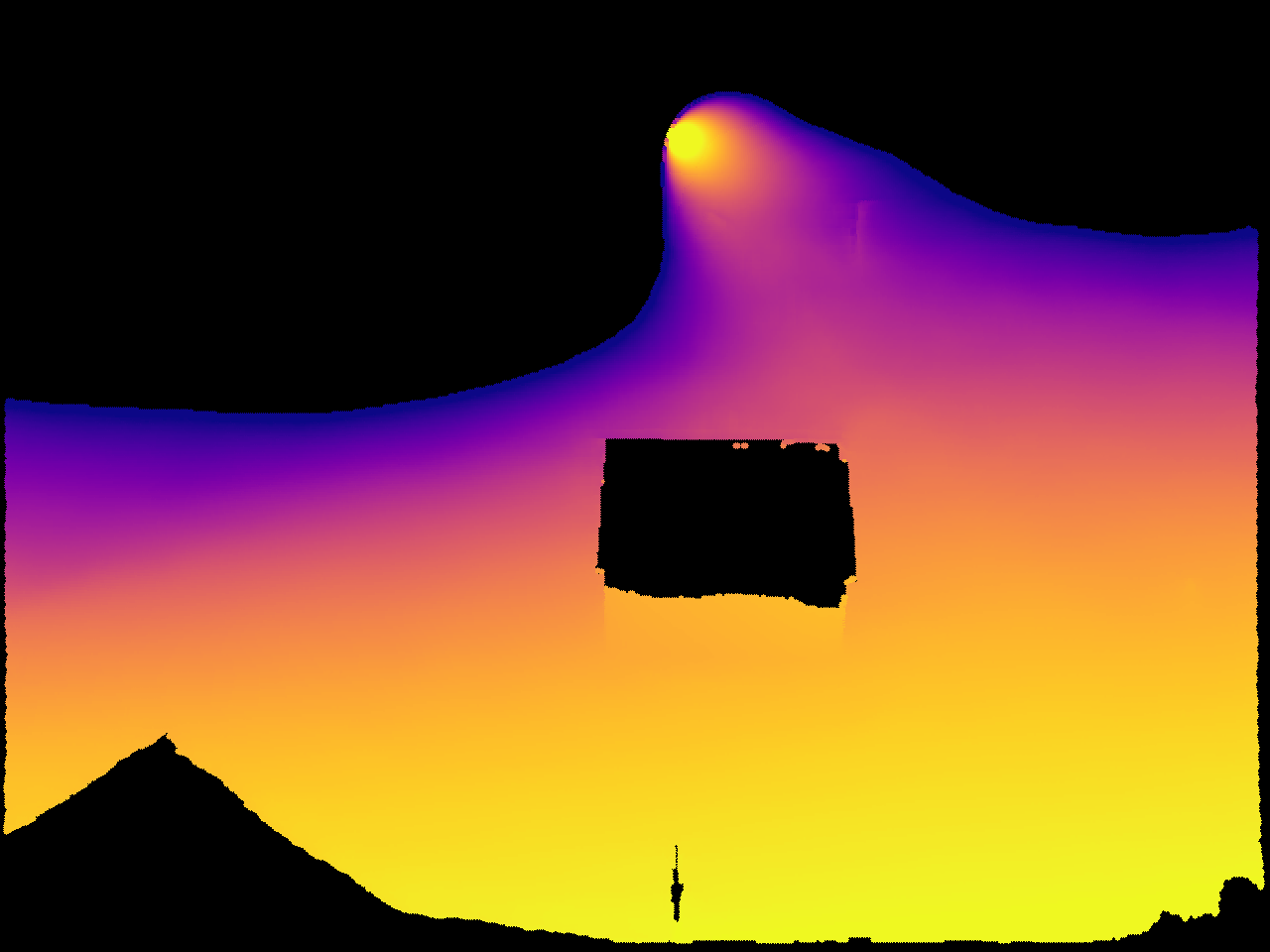}\hfill
\includegraphics[width=0.19\textwidth]{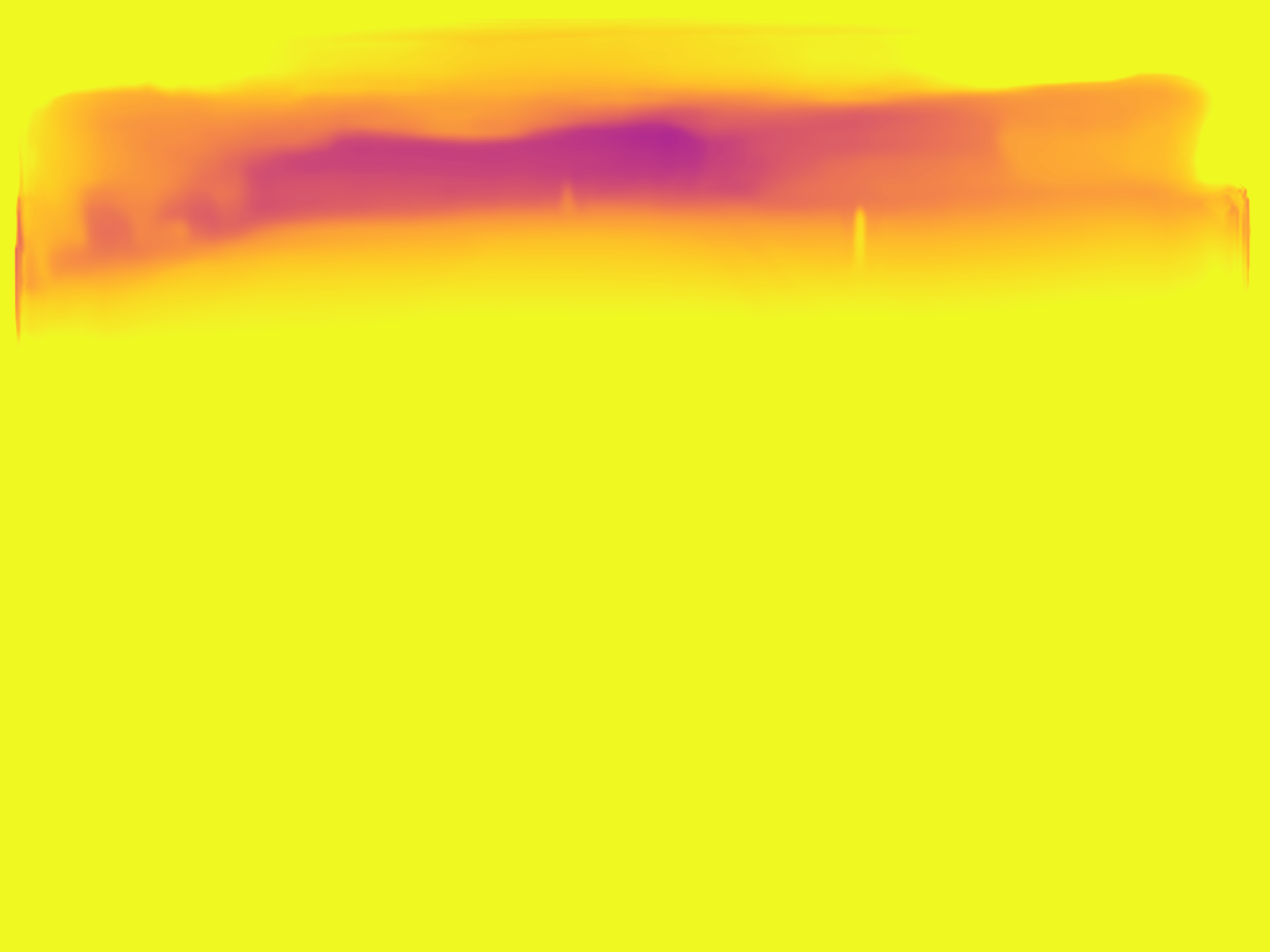}\hfill
\includegraphics[width=0.19\textwidth]{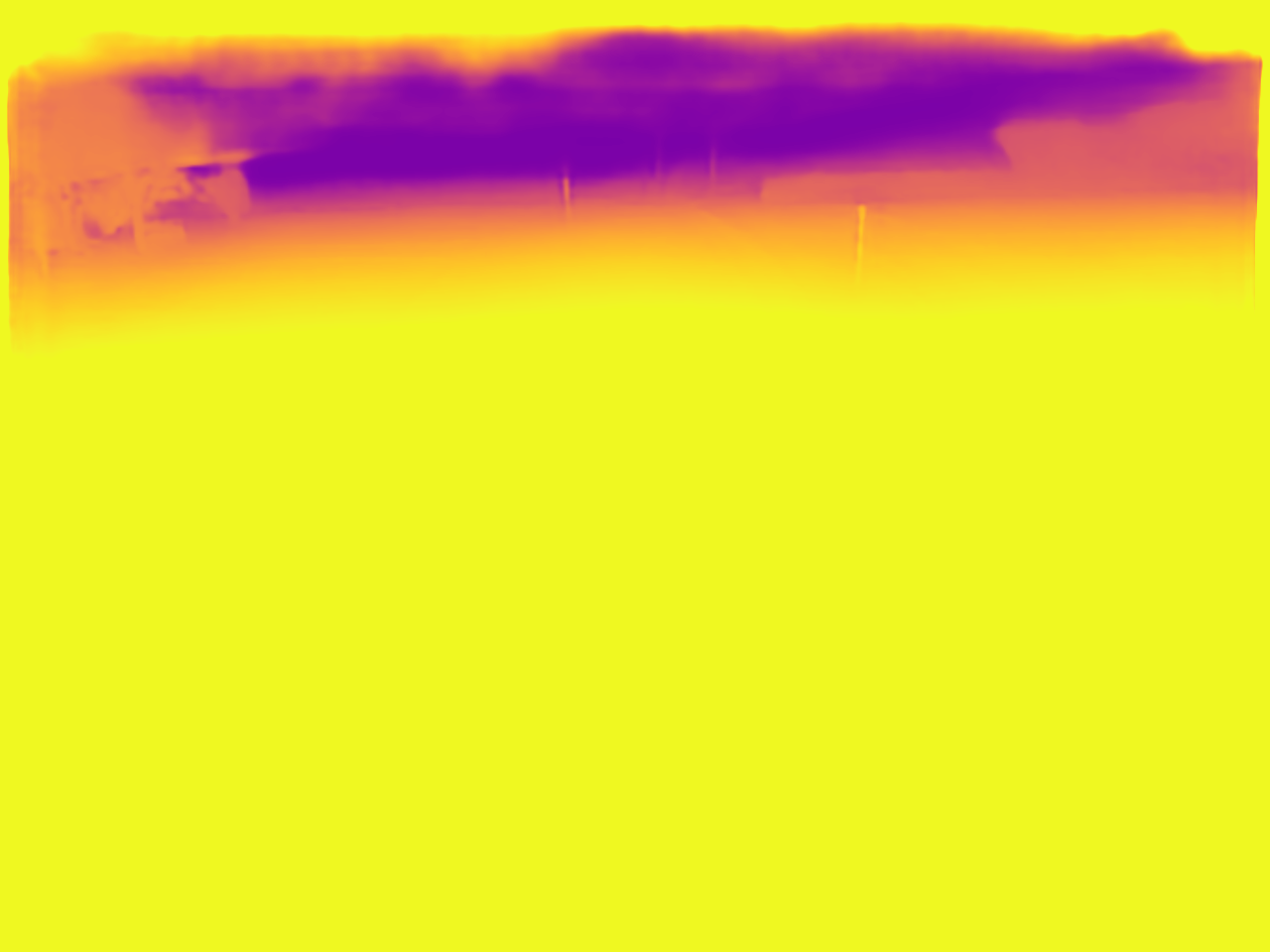}
\vspace{1pt}

% % ---------------- Row 3 ----------------

\includegraphics[width=0.19\textwidth]{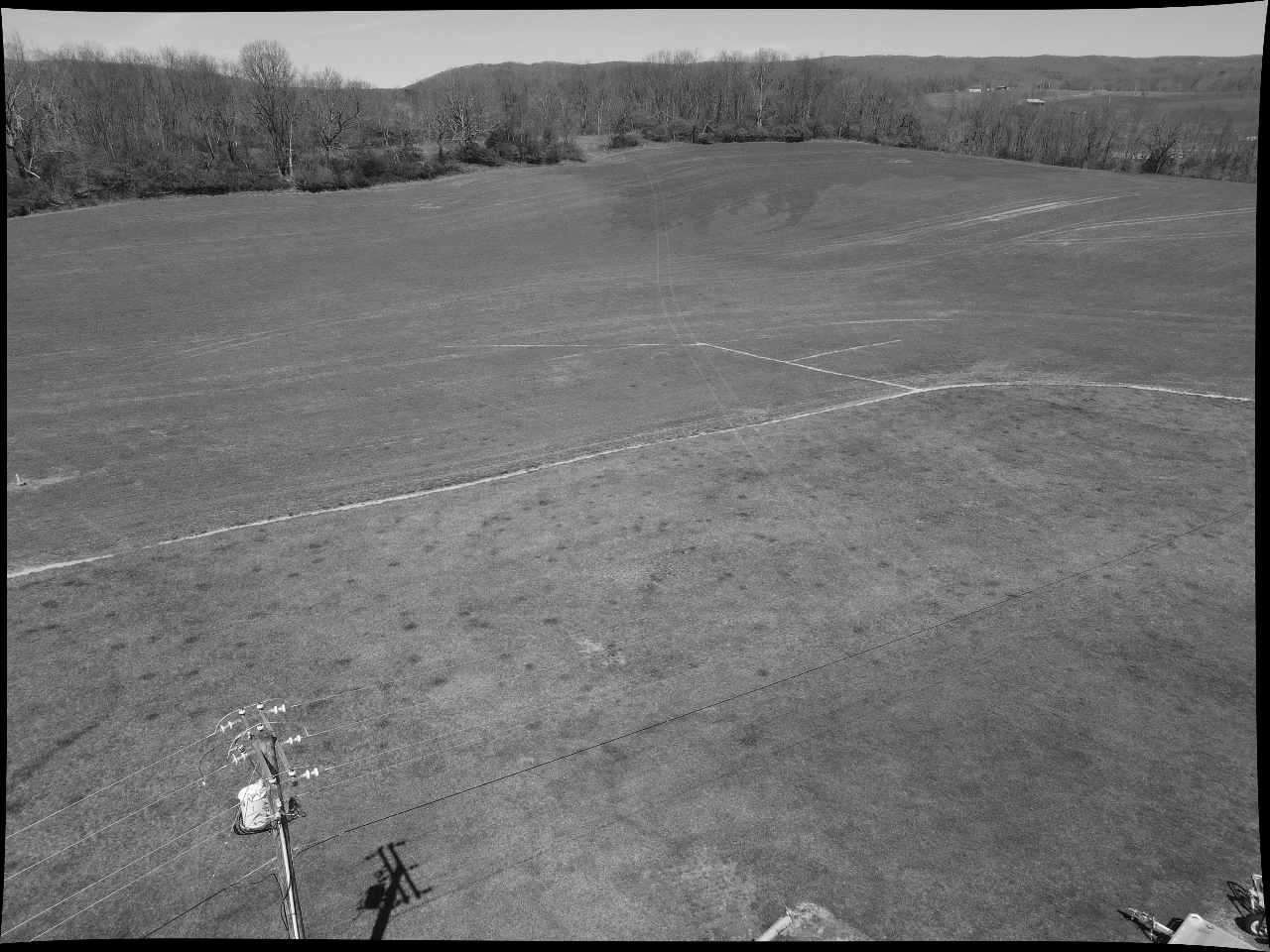}\hfill
\includegraphics[width=0.19\textwidth]{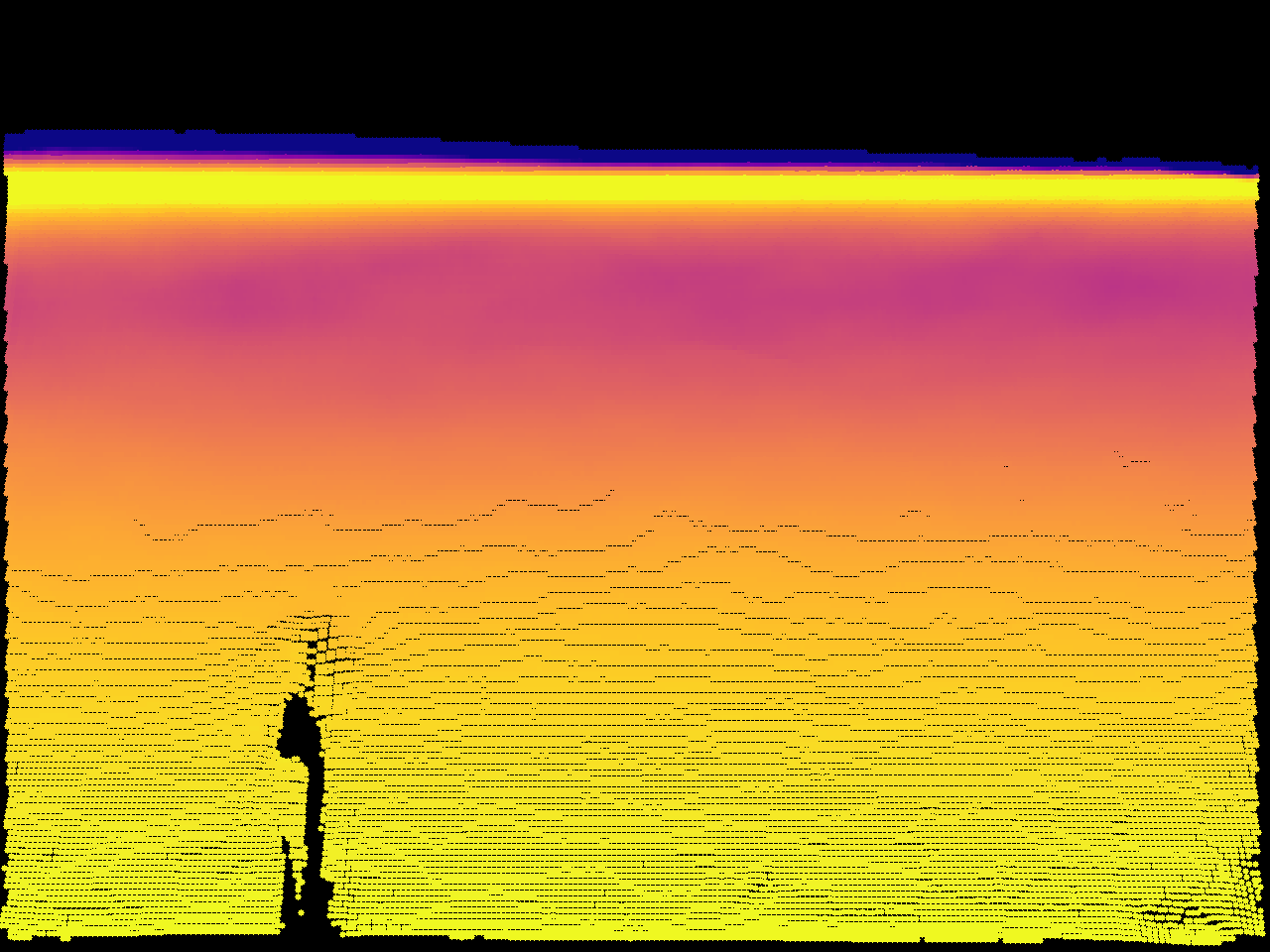}\hfill
\includegraphics[width=0.19\textwidth]{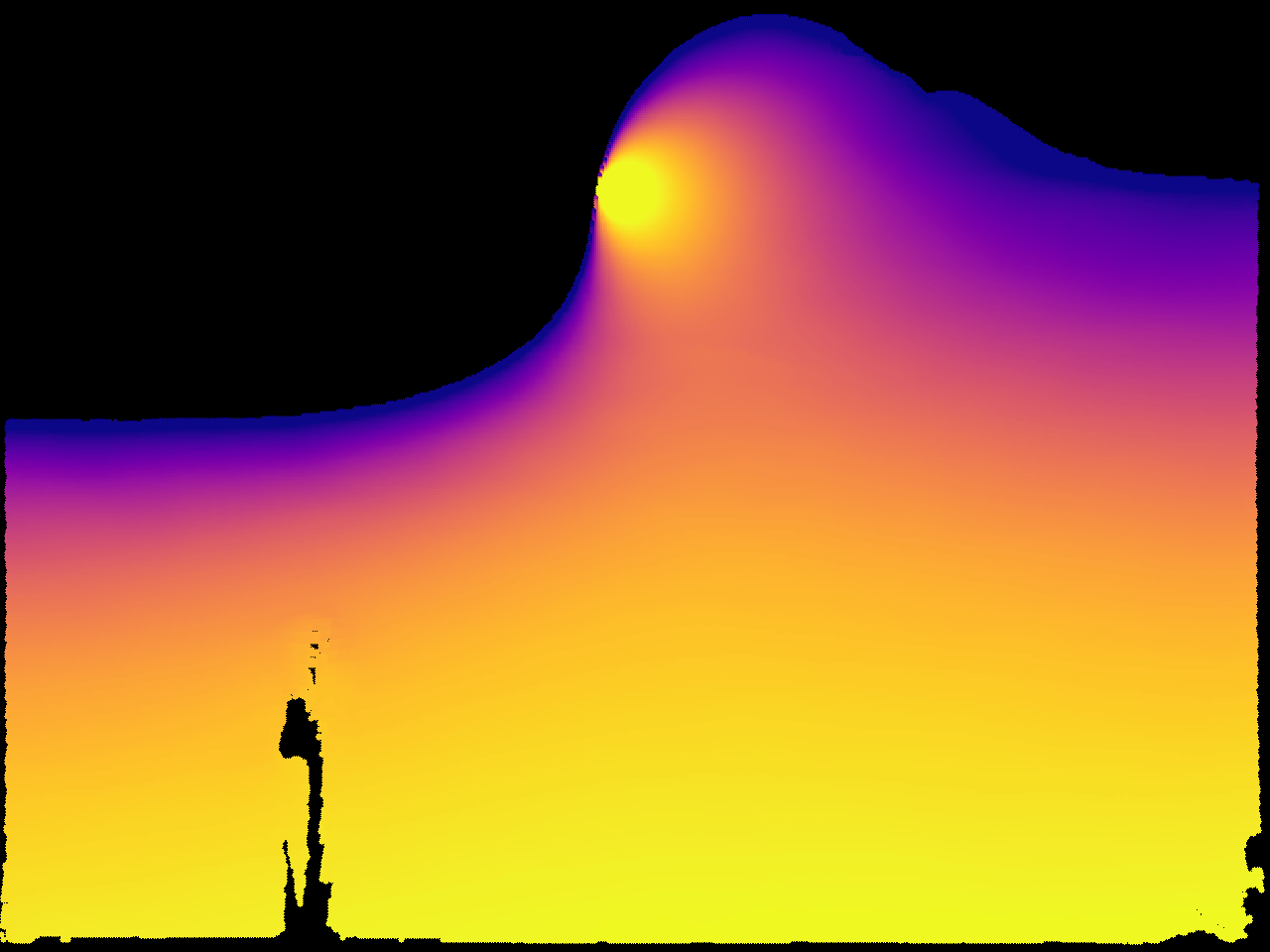}\hfill
\includegraphics[width=0.19\textwidth]{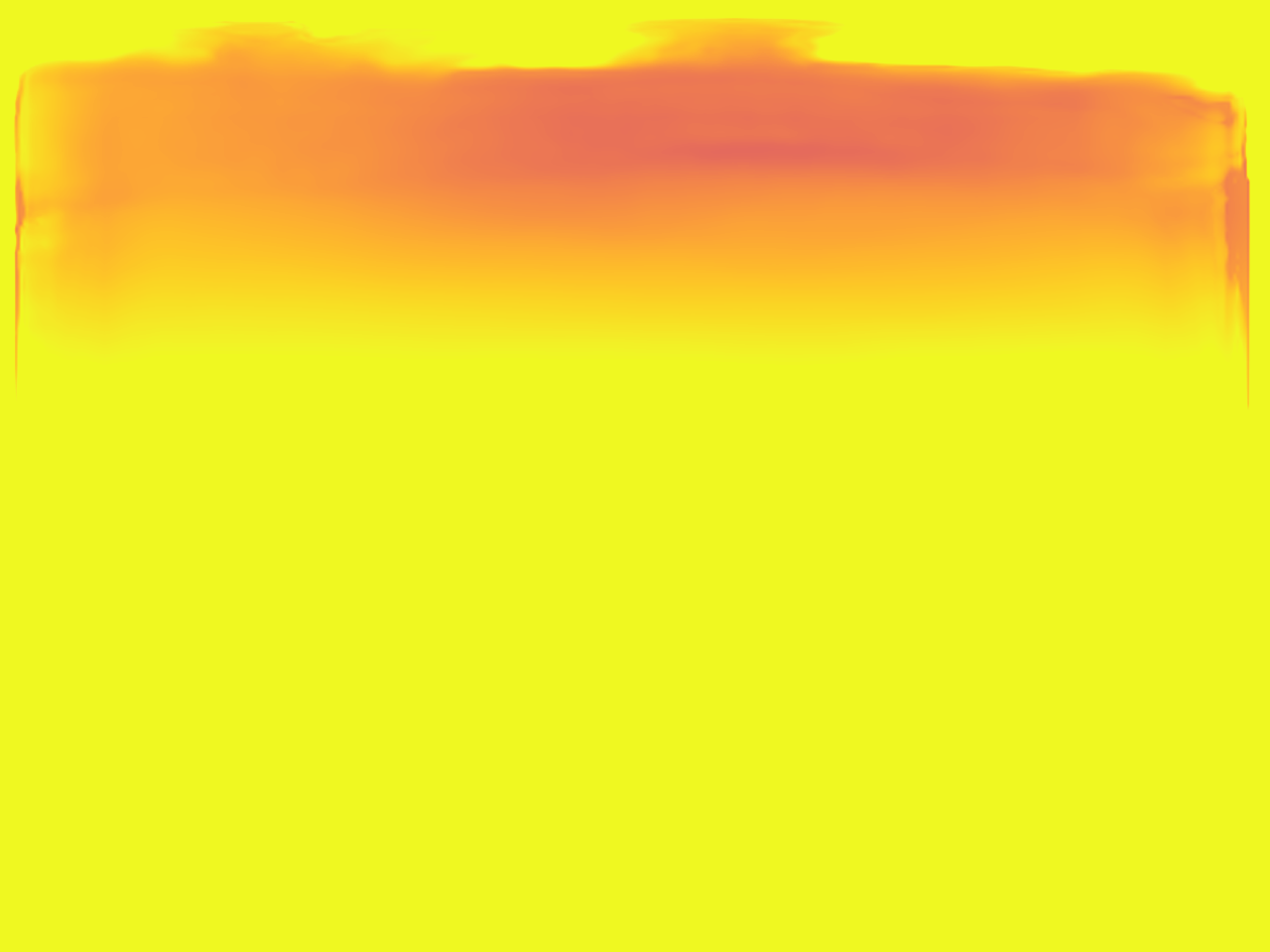}\hfill
\includegraphics[width=0.19\textwidth]{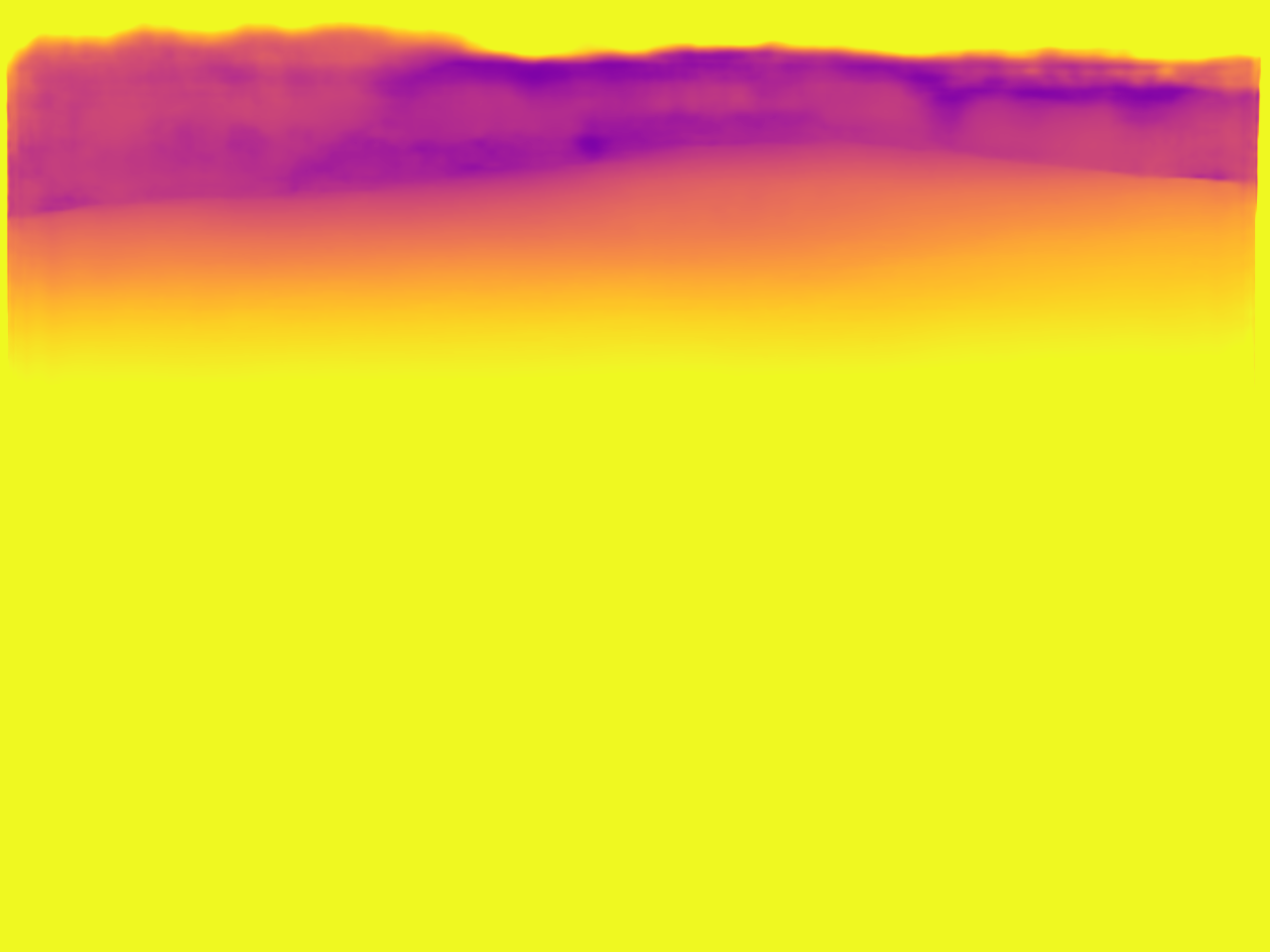}

\caption{Qualitative comparison of depth estimation results on 5 of the 8 outdoor images used for evaluation. Black regions in EpiTransfer and Triangulation indicate pixels without a valid depth estimate, either beyond the 100~m ground-truth range, unmatched in sparse correspondence, or masked as geometrically ill-conditioned for triangulation.}
\label{fig:qualitative_depth_comparison_dji}
\end{figure*}

Fig.~\ref{fig:qualitative_depth_comparison_dji} shows qualitative depth results on the outdoor dataset. Notably, the method accurately estimates depth up to a maximum range of 90 m in this scene, a range at which many existing monocular depth estimation approaches struggle to produce reliable metric predictions. A band of high-error pixels is visible near the top of the image, corresponding to points whose 3D location lies close to the plane containing the three camera centers (the trifocal plane). Our method recovers a point's location in the virtual stereo frame as the intersection of two epipolar lines: one induced by the front-to-virtual camera relation, and one induced by the second-image-to-virtual relation. For points near the trifocal plane, these two epipolar lines become nearly parallel rather than crossing at a well-conditioned angle, so a small correspondence error in either input image produces a large shift in the computed intersection point along the shared line direction. This ill-conditioned intersection is the source of the high-error band. It is a known, geometrically predictable failure mode of trifocal point transfer.

This points to a broader limitation: accurate depth estimation requires the epipole to fall outside the region of interest. Our 20\textdegree{} downward gimbal pitch keeps the epipole near the sky/horizon, above the ground-level content of practical interest. This relies on oblique viewing geometry; a forward-facing camera under primarily translational motion, as in KITTI, would place the epipole near the image center, directly overlapping the region needed for depth estimation. Our method is thus best suited to oblique or downward-looking configurations, and all evaluation sequences here were correspondingly curated to avoid near-degenerate forward motion; performance under such motion is left to future work.

Table~\ref{tab:outdoor_depth_results} reports depth accuracy on the outdoor dataset, evaluated against the USGS LiDAR ground truth described in Section~\ref{sec:setup}. Our method (EpiTransfer) substantially outperforms both learning-based baselines across all metrics, achieving an AbsRel of 0.138 compared to 0.719 and 0.744 for ZoeDepth and Depth Anything V2, respectively, with a corresponding 3.2$\times$ reduction in RMSE (10.255~m vs.\ approximately 32~m). Notably, both baselines achieve $\delta < 1.25$ accuracy of 0.000, indicating that neither method's absolute depth predictions fall within 25\% of the LiDAR ground truth for any evaluated pixel. We attribute this to the domain gap between our aerial, oblique imagery and the ground-level, forward-facing imagery on which these models are primarily trained: their predicted metric scale does not generalize to this altitude and viewing geometry, which AbsRel, RMSE, and $\delta$ directly penalize. EpiTransfer computes metric depth from known pose and geometry rather than learned priors, and is unaffected by this shift.

\begin{figure*}[t]
\centering
\small
% Column Header Titles
\begin{tabular}{ccccc}
    \makebox[0.18\textwidth]{\textbf{Input Image}} &
    \makebox[0.18\textwidth]{\textbf{EpiTransfer (Ours)}} &
    \makebox[0.18\textwidth]{\textbf{Triangulation}} &
    \makebox[0.18\textwidth]{\textbf{ZoeDepth}} &
    \makebox[0.17\textwidth]{\textbf{Depth Anything}} \\
\end{tabular}

\vspace{2pt}

% ---------------- Row 1 ----------------
\includegraphics[width=0.19\textwidth]{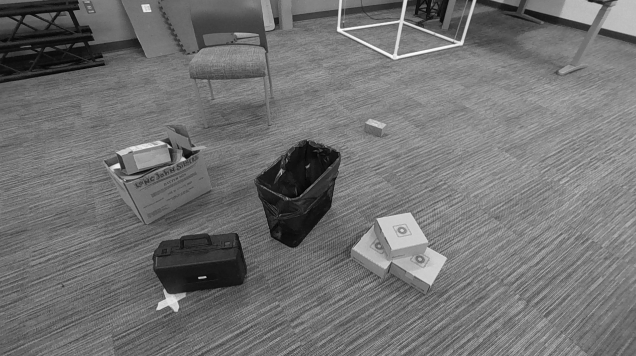}\hfill
\includegraphics[width=0.19\textwidth]{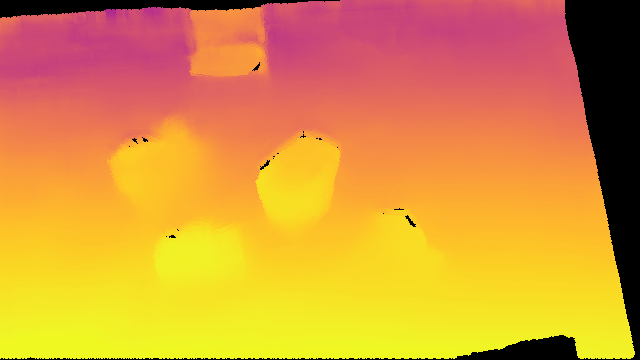}\hfill
\includegraphics[width=0.19\textwidth]{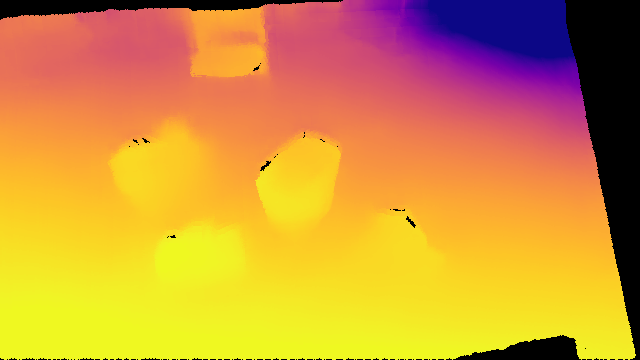}\hfill
\includegraphics[width=0.19\textwidth]{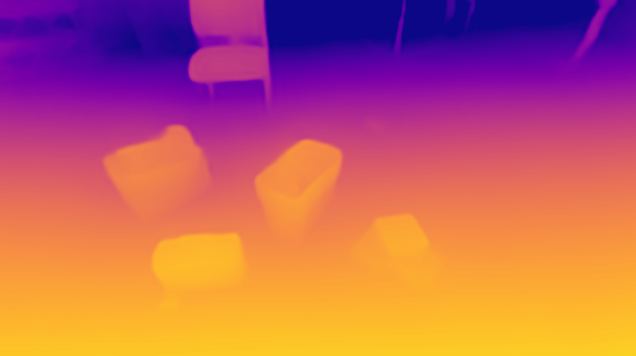}\hfill
\includegraphics[width=0.19\textwidth]{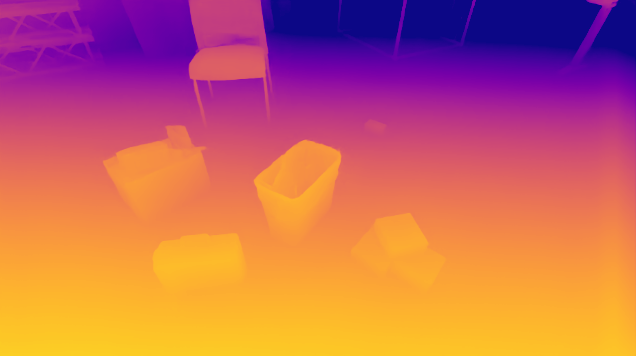}

\vspace{2pt}

% % ---------------- Row 2 ----------------
\includegraphics[width=0.19\textwidth]{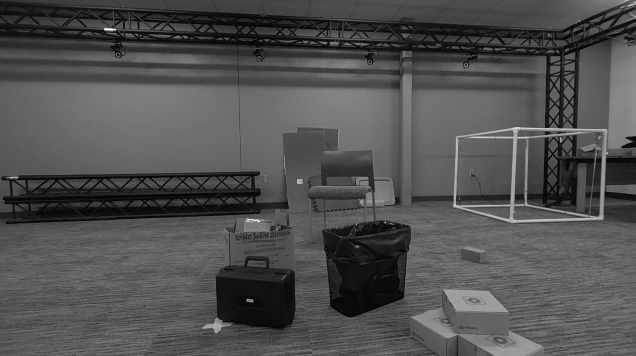}\hfill
\includegraphics[width=0.19\textwidth]{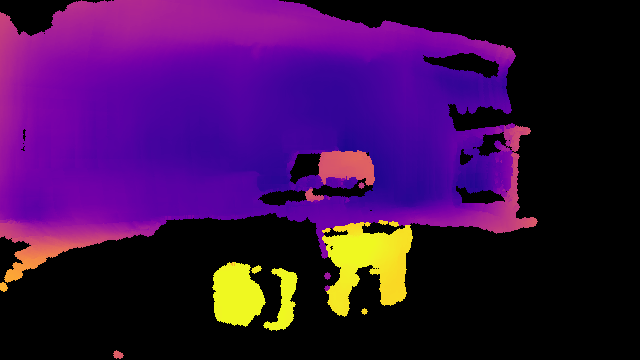}\hfill
\includegraphics[width=0.19\textwidth]{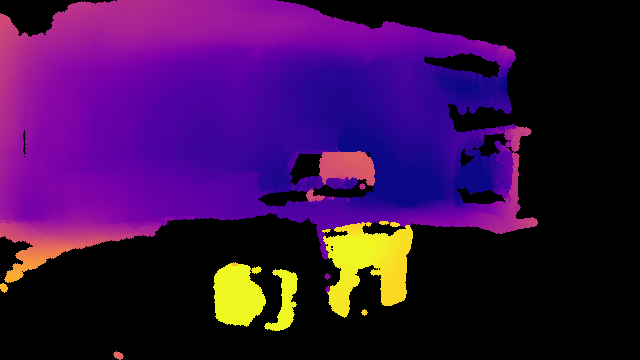}\hfill
\includegraphics[width=0.19\textwidth]{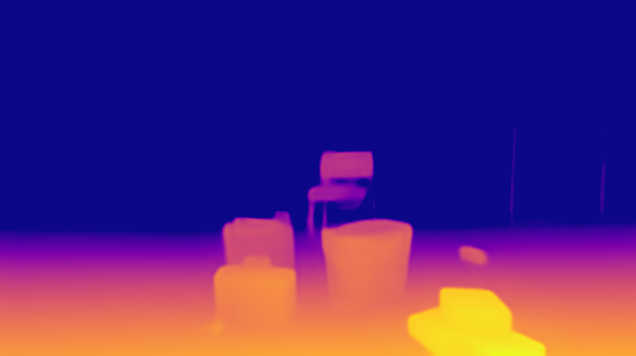}\hfill
\includegraphics[width=0.19\textwidth]{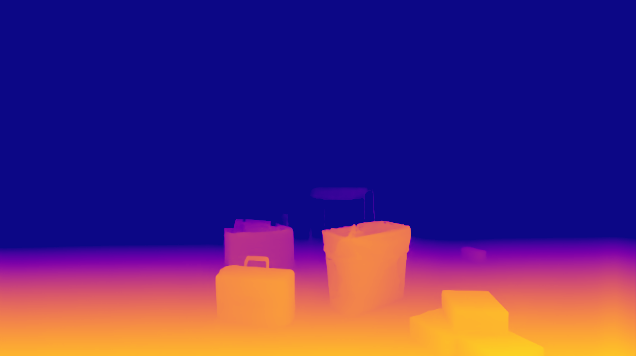}

\vspace{2pt}

% % ---------------- Row 3 ----------------
\includegraphics[width=0.19\textwidth]{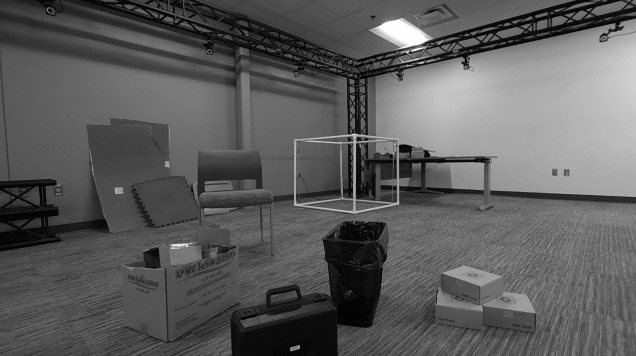}\hfill
\includegraphics[width=0.19\textwidth]{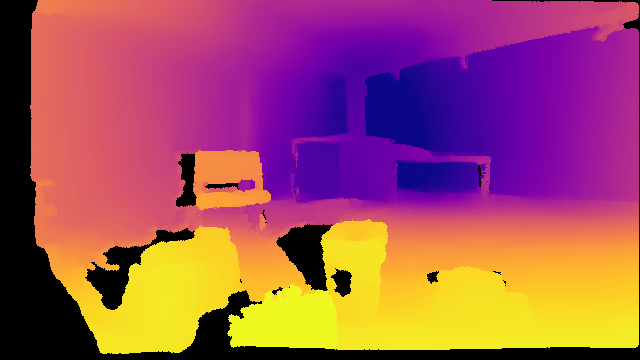}\hfill
\includegraphics[width=0.19\textwidth]{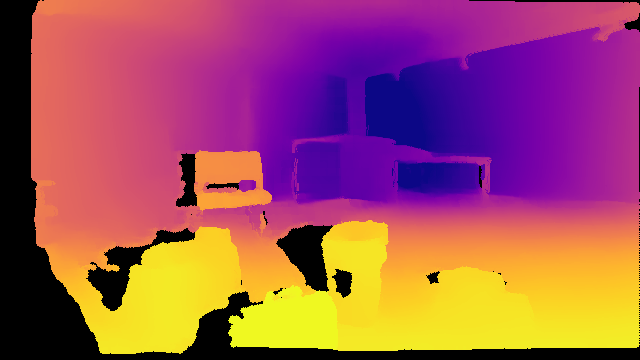}\hfill
\includegraphics[width=0.19\textwidth]{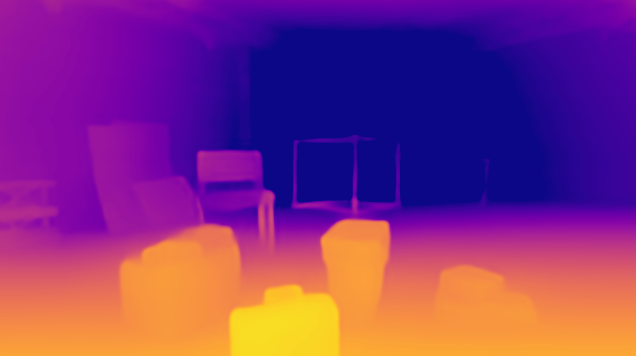}\hfill
\includegraphics[width=0.19\textwidth]{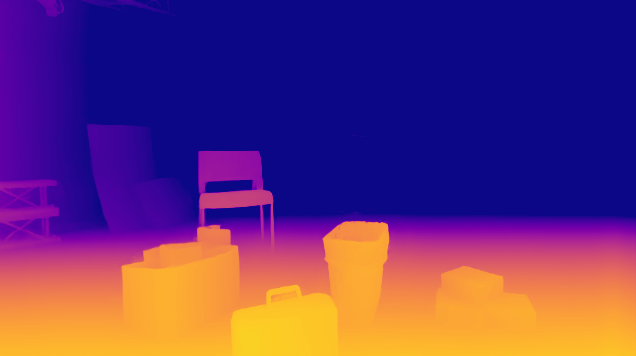}
\vspace{2pt}

\caption{Qualitative comparison of depth estimation results on the indoor dataset. Black regions in EpiTransfer and Triangulation indicate pixels without a valid depth estimate, either unmatched in sparse correspondence or masked as geometrically ill-conditioned for triangulation.}
\label{fig:qualitative_depth_comparison_indoor}
\end{figure*}

\begin{table}[htbp]
  \centering
  \caption{Depth estimation accuracy evaluated against LiDAR ground truth on 8 images of our outdoor dataset.}
  \label{tab:outdoor_depth_results}

\begin{tabular}{l@{\hspace{5pt}}c@{\hspace{5pt}}c@{\hspace{5pt}}c@{\hspace{5pt}}c}
\toprule
\textbf{Metric}
& \textbf{\makecell{EpiTransfer\\(Ours)}}
& \textbf{\makecell{Triangulation\\(DLT)}}
& \textbf{\makecell{ZoeDepth}}
& \textbf{\makecell{Depth Anything \\ V2}} \\
\midrule

AbsRel $\downarrow$
& 0.138
& 0.124
& 0.719
& 0.744 \\

RMSE (m) $\downarrow$
& 10.255
& 8.731
& 32.610
& 32.959 \\

RMSE$_{\log}$ $\downarrow$
& 0.199
& 0.174
& 1.279
& 1.381 \\

MAE (m) $\downarrow$
& 6.649
& 5.656
& 29.750
& 30.423 \\

SqRel $\downarrow$
& 1.694
& 1.350
& 21.560
& 22.579 \\

\midrule

$\delta < 1.25$ $\uparrow$
& 0.755
& 0.817
& 0.000
& 0.000 \\

$\delta < 1.25^{2}$ $\uparrow$
& 0.961
& 0.978
& 0.000
& 0.000 \\

$\delta < 1.25^{3}$ $\uparrow$
& 0.998
& 0.998
& 0.001
& 0.000 \\

\bottomrule
\end{tabular}
\end{table}
Table ~\ref{tab:outdoor_depth_results} also reports a Triangulation (DLT) baseline, which consumes
the same sparse correspondences as EpiTransfer but triangulates
directly from the two real camera centers rather than through a
synthetic virtual view. On the pixels retained by
both methods, DLT achieves marginally better aggregate accuracy
(AbsRel 0.124 vs.\ 0.138). However, black regions in Fig.\ref{fig:qualitative_depth_comparison_dji} arise
from two distinct causes: depth beyond the 100\,m LiDAR ground-truth
range is not evaluated regardless of method (e.g.\ the second row of
Fig.\ref{fig:qualitative_depth_comparison_dji}, where valid depth ends at the foreground structure), and,
separately, points near the trifocal plane are masked as
geometrically ill-conditioned for triangulation. It is
this second, method-dependent cause that differs between EpiTransfer
and DLT: in the two low-texture scenes shown in the fourth and fifth
rows of Fig.\ref{fig:qualitative_depth_comparison_dji}, EpiTransfer retained 753 and 1147 valid points,
respectively, compared to 715 and 1103 for DLT (increases of 5.3\%
and 4.0\%). This suggests that EpiTransfer's controllable virtual
baseline achieves accuracy comparable to direct triangulation while
retaining usable depth over a modestly larger portion of the scene in
geometrically challenging conditions, particularly under the
imperfect GPS/IMU pose available outdoors. Indoors, where OptiTrack
pose is far more accurate, the two methods produce nearly
indistinguishable masks and depth maps (Fig.\ref{fig:qualitative_depth_comparison_indoor}).

\subsection{Indoor}

Table~\ref{tab:depth_results} reports results on 7 representative indoor frames, evaluated against the Velodyne LiDAR ground truth. EpiTransfer achieves an AbsRel of 0.092 and $\delta < 1.25$ of 0.940, comparable to DLT, which achieves a marginally better AbsRel of 0.073 and $\delta < 1.25$ of 0.943 under the accurate OptiTrack pose available indoors. Both geometric methods substantially outperform the learning-based baselines: ZoeDepth achieves a more competitive but still considerably worse AbsRel of 0.225 ($\delta < 1.25 = 0.601$), while Depth Anything V2 degrades further, with an AbsRel of 0.570 and $\delta < 1.25$ of only 0.223. This indicates that even in a domain closer to these models' training distribution than our outdoor sequences, learning-based metric depth remains less reliable than geometry-based estimation when accurate relative pose is available, consistent with the domain-shift limitations discussed in Section~\ref{sec:learning-based}, even in scenes with the low-texture regions described in Section~\ref{sec:data_collection}.

\begin{table}[htbp]
  \centering
  \caption{Depth estimation accuracy evaluated against LiDAR ground truth on 7 images of our indoor dataset.}
  \label{tab:depth_results}
\begin{tabular}{l@{\hspace{5pt}}c@{\hspace{5pt}}c@{\hspace{5pt}}c@{\hspace{5pt}}c}
\toprule
\textbf{Metric}
& \textbf{\makecell{EpiTransfer\\(Ours)}}
& \textbf{\makecell{Triangulation \\(DLT)}}
& \textbf{\makecell{ZoeDepth}}
& \textbf{\makecell{Depth \\Anything V2}} \\
\midrule

AbsRel $\downarrow$
& 0.092
& 0.073
& 0.225
& 0.570 \\

RMSE (m) $\downarrow$
& 0.469
& 0.401
& 0.980
& 2.964 \\

RMSE$_{\log}$ $\downarrow$
& 0.118
& 0.102
& 0.235
& 0.482 \\

MAE (m) $\downarrow$
& 0.371
& 0.303
& 0.841
& 2.520 \\

SqRel $\downarrow$
& 0.052
& 0.037
& 0.266
& 1.859 \\

\midrule

$\delta < 1.25$ $\uparrow$
& 0.940
& 0.943
& 0.601
& 0.223 \\

$\delta < 1.25^{2}$ $\uparrow$
& 0.998
& 0.998
& 0.971
& 0.337 \\

$\delta < 1.25^{3}$ $\uparrow$
& 1.000
& 1.000
& 1.000
& 0.995 \\

\bottomrule
\end{tabular}
\end{table}

\subsection{Sensitivity Analysis}

\begin{figure}[t]
  \centering
  \begin{subfigure}{0.48\columnwidth}
    \includegraphics[width=\linewidth]{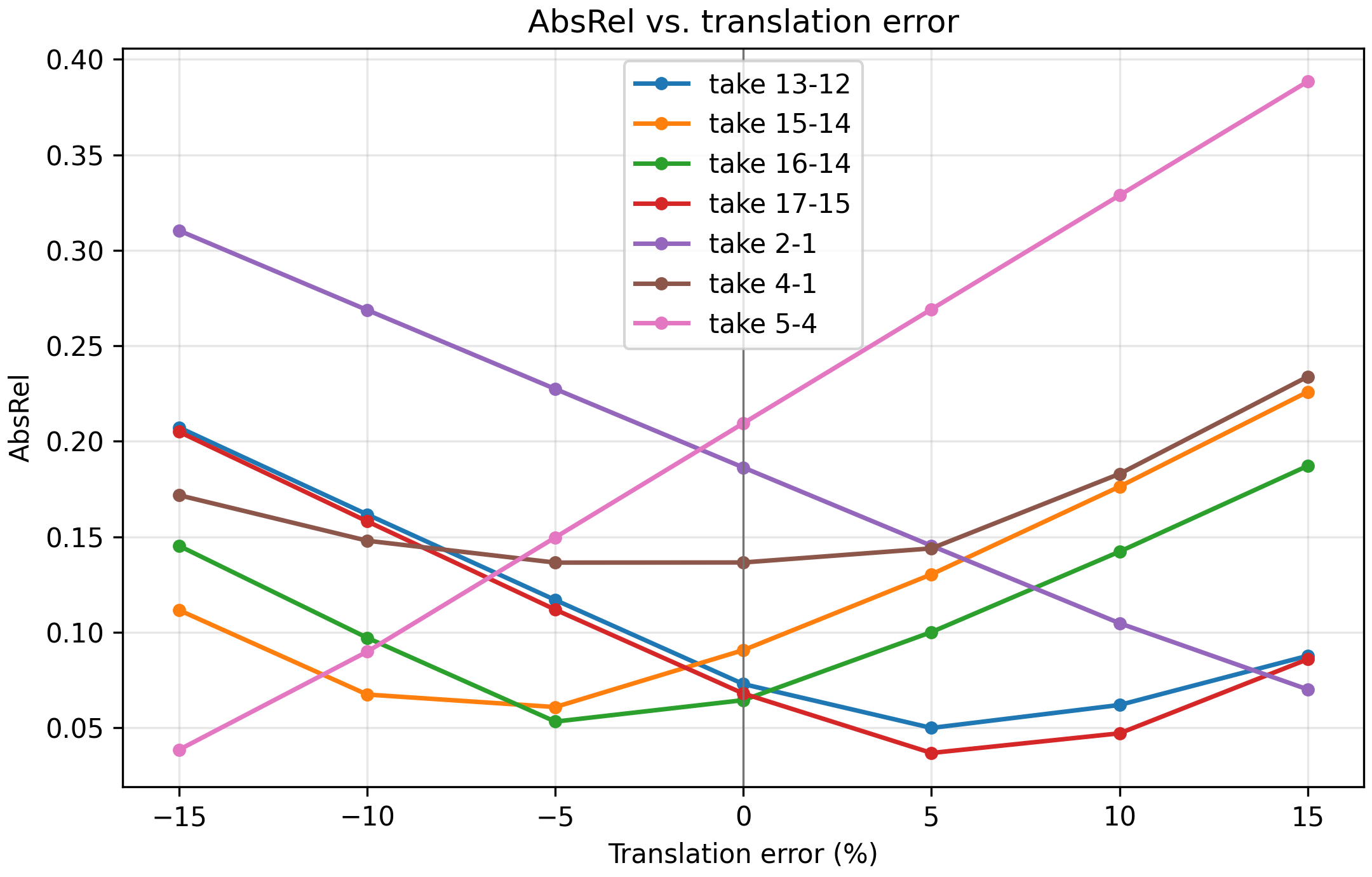}
    \caption{Translation error}
    \label{fig:sens_translation}
  \end{subfigure}
  \hfill
  \begin{subfigure}{0.48\columnwidth}
    \includegraphics[width=\linewidth]{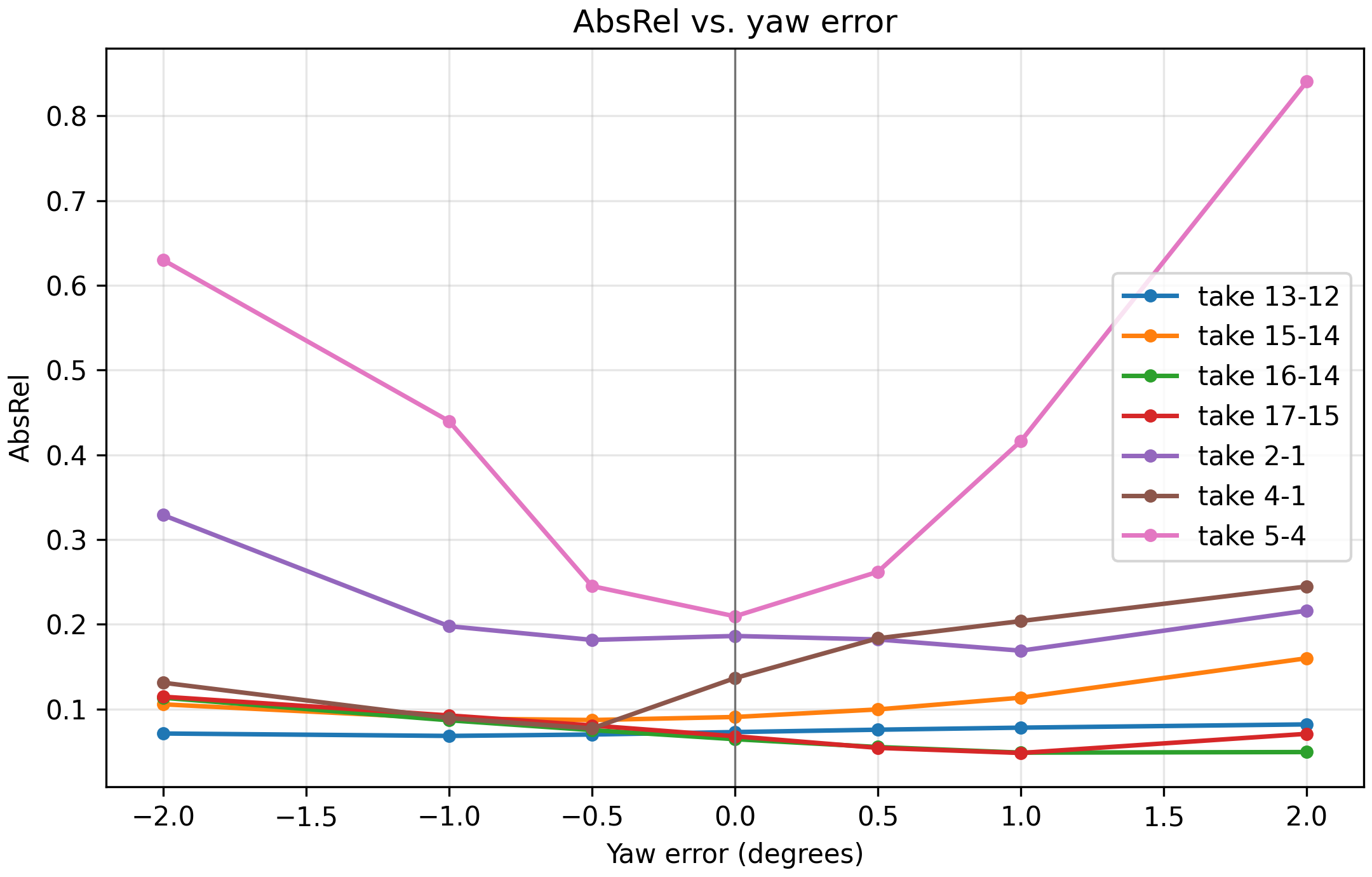}
    \caption{Yaw error}
    \label{fig:sens_yaw}
  \end{subfigure}

  \vspace{2pt}

  \begin{subfigure}{\columnwidth}
    \includegraphics[width=\linewidth]{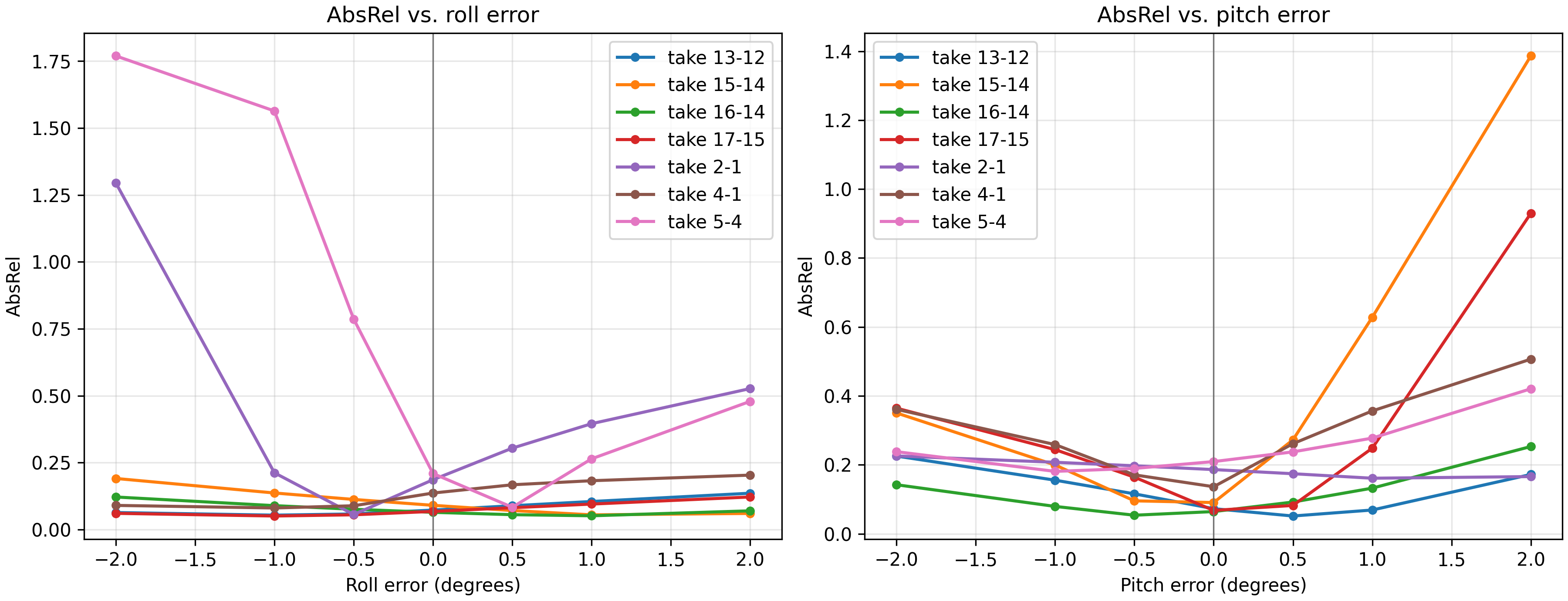}
    % \caption{Roll and pitch error}
    \label{fig:sens_rollpitch}
  \end{subfigure}

  \caption{AbsRel vs. injected pose error across 7 indoor takes. Takes 2-1 and 5-4 (which include perturbations placing the epipole inside the image) show markedly larger, non-symmetric error, consistent with the trifocal-plane degeneracy of Section \ref{sec:res_outdoor}; the remaining 5 takes show bounded, approximately symmetric degradation around zero injected error.}
  \label{fig:sensitivity}
\end{figure}

Fig.~\ref{fig:sensitivity} reports AbsRel vs. injected translation, yaw, roll, and pitch error on 7 indoor takes, using OptiTrack pose as a clean zero-noise reference. Five takes show bounded, roughly symmetric degradation around zero injected error. Takes 2-1 and 5-4 instead show substantially larger, near-monotonic error; in both cases, several tested perturbations place the epipole inside the image, reproducing the trifocal-plane degeneracy of Section~V-A. This confirms that pose-error sensitivity is governed less by a uniform error bound than by proximity to this geometric degeneracy, which can be checked directly from the candidate pose to flag unreliable estimates in practice.

\section{Conclusion}
\label{sec:conclusion}
In this paper, we presented EpiTransfer, a training-free geometric
method for monocular sparse depth estimation via epipolar transfer.
By synthesizing a virtual stereo pair with a freely chosen baseline
from known camera poses, our approach avoids dedicated stereo
hardware and multi-frame optimization while achieving accuracy comparable to direct two-view triangulation
and substantially better accuracy than off-the-shelf, pretrained
learning-based monocular depth baselines not trained for this
domain, across both indoor OptiTrack and outdoor drone-flight
evaluations against LiDAR ground truth, with a validated range of approximately
90\,m outdoors. Qualitative results further suggest that EpiTransfer
retains valid depth over a larger portion of challenging,
low-texture scenes than direct triangulation, particularly under
the imperfect pose available outdoors. With its minimal
computational overhead (10\,Hz on embedded hardware), this
framework provides an efficient, training-free sparse prior for
depth completion, obstacle avoidance, and real-time robotic
navigation. Future work will complete flight testing of the onboard XFeat-based pipeline against LiDAR ground truth on the hexacopter platform and extend evaluation across a broader range of scenes and flight conditions.

\section*{ACKNOWLEDGMENT}

Generative AI (Claude, Anthropic) assisted in drafting and revising
the Introduction, Related Work, Abstract, and Conclusion, and in
regenerating Figs.~1--2 for legibility; all content was reviewed and
verified by the authors.

% \addtolength{\textheight}{-12cm}   % This command serves to balance the column lengths
                                  % on the last page of the document manually. It shortens
                                  % the textheight of the last page by a suitable amount.
                                  % This command does not take effect until the next page
                                  % so it should come on the page before the last. Make
                                  % sure that you do not shorten the textheight too much.

%%%%%%%%%%%%%%%%%%%%%%%%%%%%%%%%%%%%%%%%%%%%%%%%%%%%%%%%%%%%%%%%%%%%%%%%%%%%%%%%

%%%%%%%%%%%%%%%%%%%%%%%%%%%%%%%%%%%%%%%%%%%%%%%%%%%%%%%%%%%%%%%%%%%%%%%%%%%%%%%%
\bibliographystyle{IEEEtran}

\bibliography{references}

\end{document}